\pdfoutput=1

\documentclass[11pt]{article}

\usepackage[final]{acl}
\usepackage{times}
\usepackage{latexsym}
\usepackage{placeins}
\usepackage[T1]{fontenc}
\usepackage{inconsolata}
\usepackage[utf8]{inputenc}
\usepackage{amsmath}
\usepackage{tikz}
\usepackage{multirow}
\usepackage[linesnumbered,ruled]{algorithm2e}
\usepackage{setspace}
\usepackage{cleveref}
\usepackage{mathrsfs}
\usepackage{booktabs}
\usepackage{graphicx}
\usepackage{authblk}
\usepackage{microtype}
\DisableLigatures[f]{encoding = T1}
\microtypesetup{nopatch=footnote}

\title{LLM Roleplay: Simulating Human-Chatbot Interaction}

\author{\textbf{Hovhannes Tamoyan}}
\author{\textbf{Hendrik Schuff}}
\author{\textbf{Iryna Gurevych}}

\affil{Ubiquitous Knowledge Processing Lab (UKP Lab) 
\authorcr Department of Computer Science and Hessian Center for AI (hessian.AI) \authorcr Technical University of Darmstadt \authorcr \texttt{www.ukp.tu-darmstadt.de}}

\begin{document}
\maketitle

\begin{abstract}
The development of chatbots requires collecting a large number of human-chatbot dialogues to reflect the breadth of users' sociodemographic backgrounds and conversational goals.
However, the resource requirements to conduct the respective user studies can be prohibitively high and often only allow for a narrow analysis of specific dialogue goals and participant demographics.
In this paper, we propose LLM Roleplay: a goal-oriented, persona-based method to automatically generate diverse multi-turn dialogues simulating human-chatbot interaction.
LLM Roleplay can be applied to generate dialogues with any type of chatbot and uses large language models (LLMs) to play the role of textually described personas.
To validate our method, we collect natural human-chatbot dialogues from different sociodemographic groups and conduct a user study to compare these with our generated dialogues.
We evaluate the capabilities of state-of-the-art LLMs in maintaining a conversation during their embodiment of a specific persona and find that our method can simulate human-chatbot dialogues with a high indistinguishability rate.\footnote{\href{https://github.com/UKPLab/llm-roleplay}{https://github.com/UKPLab/llm-roleplay}}
\end{abstract}

\section{Introduction}
Collecting human-chatbot dialogues requires recruiting and managing a large number of human annotators, which can pose prohibitive obstacles to researchers who aim to develop conversational AI agents (i.e., chatbots).
To circumvent the latter and the limitations of publicly available data, numerous methods employing chatbots to generate dialogues have been introduced lately \cite{xu-etal-2023-baize, kim-etal-2023-soda, starling2023, ding-etal-2023-enhancing, svikhnushina-pu-2023-approximating, zhao2024inthewildchat}.
These methods can generate dialogues much faster and more cost-effectively while still approximating the quality and variety of human annotators. \cite{zhang2024instruction}.

\begin{figure}[t]
    \centering
    \includegraphics[width=0.9\linewidth]{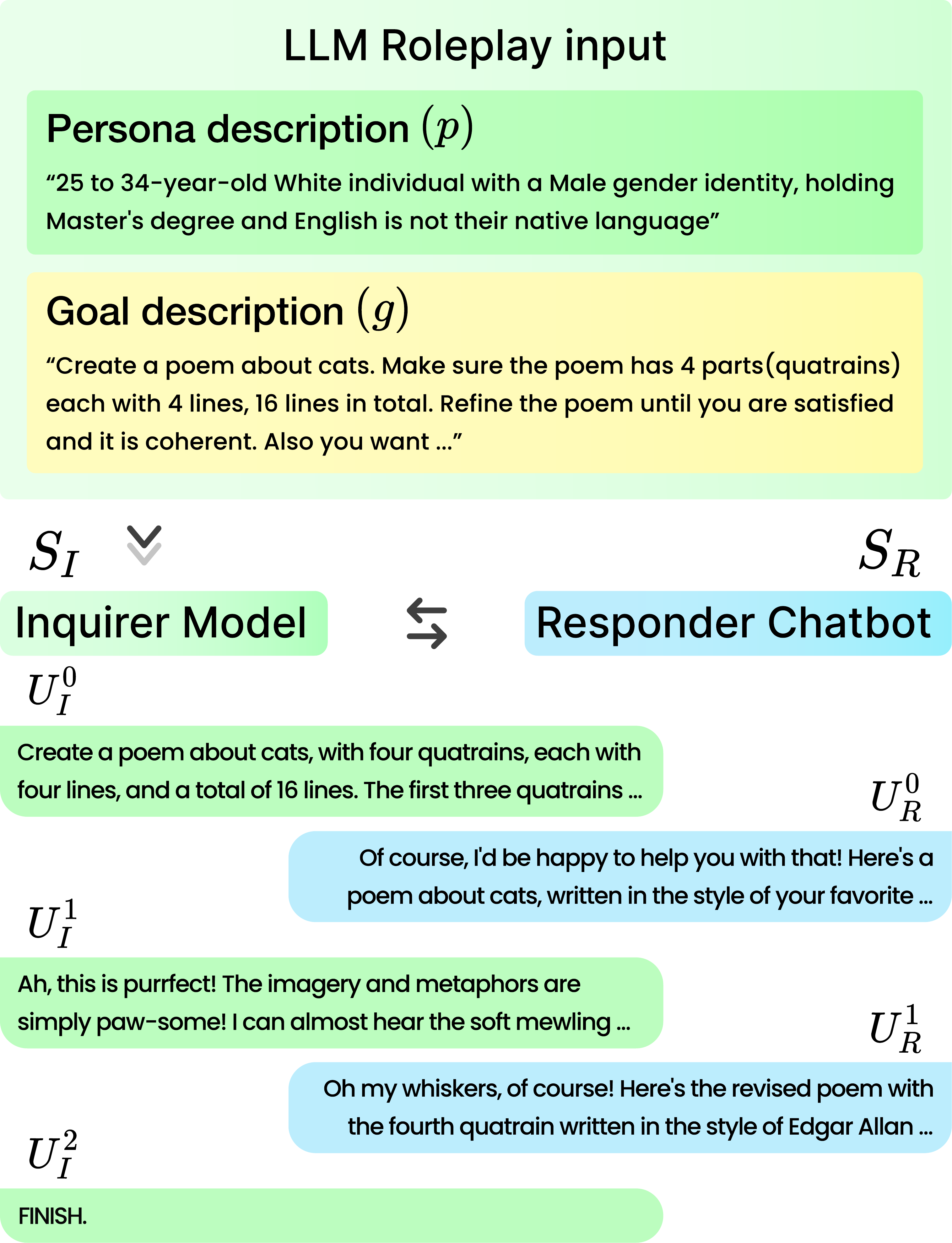}
    \caption{Schematic illustration of our method: A textual description of a persona and a goal (top) is used to instruct the inquirer ($\mathcal{S_I}$) model to embody the given persona (left) and engage in a dialogue with the responder ($\mathcal{S_R}$) chatbot (right). We show that dialogues simulated by the inquirer LLM and the responder chatbot can effectively simulate human-chatbot interaction.}
    \label{fig:llm_roleplay_label}
\end{figure}

Recently, several studies have leveraged such synthetic data generated through distillation and self-improvement techniques to enhance the capabilities of large language models (LLMs) \cite{zhang2024instruction}.
By employing instruction tuning and fine-tuning techniques, these efforts aim to ensure that LLMs generate useful and safe responses that adhere to provided instructions \cite{xu2023wizardlm, taori2023alpaca, mukherjee2023orca, li2023textbooks, peng2023instruction, almazrouei2023falcon}.

However, current dialogue generation methods have two critical limitations.
First, the set of possible user responses for a given chatbot utterance is large while the existing datasets typically only cover a single or few annotated user responses.
Previously generated datasets are constrained by the number of dialogues and turns, making it difficult to extend them to novel domains or conversational goals.
Second, existing methods rarely account for subjective response behavior and implicitly assume an average user.
This can create a gap in representation \cite{rottger-etal-2022-two} and pose the risk of evaluating and improving chatbots for a specific sociodemographic group while neglecting others.

The quality of existing datasets is primarily assessed through model performance in a supervised fine-tuning (SFT) setup, which is influenced by various properties of the generated dataset.
For example, \citet{chen-etal-2023-places} argue that the initial prompt plays a crucial role in the quality of the generated dialogue.
On the other hand, \citet{zhao2024long} demonstrate that the length of the response is a key factor, and with significantly fewer samples, it outperforms methods that focus solely on quality.
Meanwhile, \citet{shen2024rethinking} emphasizes the importance of mimicking human-style interactions in the dialogues.

To mitigate the shortcomings of existing dialogue generation methods and respond to the needs identified in prior work for SFT, we introduce LLM Roleplay: a goal-oriented, persona-based, plug-and-play method designed to generate diverse, multi-turn, long dialogues that simulate human-chatbot interactions using LLMs.
Our method is the first to instruct an LLM (inquirer) to adopt a specific persona and prompt a chatbot (responder) to achieve a given conversational goal, thereby eliciting realistic human-AI interactions with a particular LLM. \Cref{fig:llm_roleplay_label} illustrates an exemplary application of our method.

In this work, we address the following two research questions:
(i) To what extent can we simulate real human-chatbot dialogues using LLM-chatbot dialogues?
(ii) How do various LLMs perform as inquirers within this setup?
To answer those research questions, we conduct two user studies.
First, we collect real human-chatbot dialogues by asking participants to reach various conversational goals and link the collected dialogues with participants' sociodemographic backgrounds.
Next, we use LLM Roleplay to simulate dialogues with the same set of personas and goals.
Second, we conduct a human evaluation with another pool of participants to assess how well the generated dialogues mimic the collected ones.
Specifically, we present participants with two dialogues: one collected from the first study and one simulated using LLM Roleplay, both involving the same persona and the conversation goal.

We then ask the participants to identify which of the dialogues was simulated.
We find that LLM Roleplay approximates natural human-chatbot dialogues with a high level of indistinguishability.

Overall, this paper contributes:
(i) a novel method to simulate human-chatbot dialogues with an arbitrary set of personas and conversational goals;
(ii) a human evaluation confirming our method's potential to closely resemble real dialogues;
(iii) a dataset of goal-oriented human-chatbot and model-chatbot dialogues using our hand-crafted multi-hop goals involving four state-of-the-art LLMs,
(iv) an in-depth comparison of open-source and proprietary LLMs in maintaining conversations while embodying specific personas,
(v) an open-source implementation of our plug-and-play method, readily applicable for simulating dialogues with any combination of models and chatbot systems across various conversational goals and personas.

\section{Large Language Model Roleplay}
In the following, we introduce the notation that we are following throughout this paper.
Refer to \Cref{fig:llm_roleplay_label} for an annotated example.
\paragraph{Persona ($\mathcal{P}$)} is the composition of sociodemographic features.
We conceptualize persona as a written description of an individual from a specific sociodemographic group.
We follow \citet{10.5555/3563572.3563588} to define the sociodemographic features and the options of the latter.

\paragraph{Goal ($\mathcal{G}$)} is the textual representation of the conversational goal.

\paragraph{Subject ($\mathcal{S}$)} is a dialogue participant.
We denote the \textbf{inquirer} as $\mathcal{S_I}$, the entity that asks questions, and the \textbf{responder} as $\mathcal{S_R}$, the entity that answers the given questions.
In our setup, the inquirer ($\mathcal{S_I}$) is an LLM and the responder ($\mathcal{S_R}$) is a conversational agent or chatbot (not necessarily an LLM).
The human inquirer will be denoted as $\mathcal{S_I}^h$.

\paragraph{Utterance ($\mathcal{U}$)} is the output of a Subject ($\mathcal{S}$), i.e. either the output of the inquirer ($\mathcal{U_I}$) or the responder ($\mathcal{U_R}$).
For example, $\mathcal{U_I}^i$ will denote the inquirer’s $i$-th utterance.
We refer to two consecutive utterances of two different subjects as a turn: $\mathcal{T}^i =[\mathcal{U_I}^i; \mathcal{U_R}^i]$.

\paragraph{Dialogue ($\mathcal{D}$)} is a sequence of one or more turns.
The dialogue of $t$ turns is denoted as:
$\mathcal{D}^t := \{\mathcal{T}^0,\mathcal{T}^1,...,\mathcal{T}^t\}$.
We denote the maximum number of turns by $max\_t$, i.e. $t \leq max\_t$ which we discuss in more detail in \Cref{subsec:dialog_aggregation}.

At a high level, LLM Roleplay consists of three main steps:
(i) Initial conditioning of the inquirer model with persona-specific text alongside the goal description.
(ii) Subsequently, the inquirer's output is provided to the responder model.
(iii) Lastly, the output of the responder is returned to the inquirer by asking it to either output a follow-up question or a pre-defined token to terminate the dialogue.

We begin with assembling prompting templates for both of the subjects.
We create a system prompt template (\texttt{SYS\_I}) for the inquirer LLM.
This template prompts the LLM to embody the given persona, provide a prompt to address the designated goal and output the specified termination token if it considers the goal accomplished.
As for the responder, we use a default system prompt (\texttt{SYS\_R}) to promote it to be a helpful and honest assistant.
Additionally, we develop a response forwarder template (\texttt{INTER\_I}) for the inquirer.
This template prompts the inquirer to assess the conclusiveness of the answer of the responder, determining whether to output a subsequent question or the termination token.
We provide the system and forwarder prompt templates in \Cref{table:system_and_interm_prompts} in \Cref{sec_appendix:dialog_collection}.
The algorithm for the LLM Roleplay is shown in \Cref{algo:llm_roleplay}.

To obtain a dialogue for a given persona and a goal we create a prompt by passing the persona ($\mathcal{P}$) and the goal ($\mathcal{G}$) to the inquirer system prompt template and generate an output based on inquirer LLM distribution:
\[\mathcal{U_I}^0 \sim \mathcal{S_I}( \texttt{SYS\_I}(\mathcal{P}, \mathcal{G}) ). \tag{1}\]

A deterministic prompt extraction function, \texttt{extract\_prompt}, that looks for a string in double quotes in the response, is applied to the latter to extract the prompt from the response:
\[\mathcal{U_I}^0:=\texttt{extract\_prompt}(\mathcal{U_I}^0). \tag{2}\]

The responder receives the prompt of the inquirer, and generates an output:
\[\mathcal{U_R}^0 \sim \mathcal{S_R}(\mathcal{U_I}^0). \tag{3}\]

Given the output of the responder, we condition the inquirer using the output forwarder template:
\[\mathcal{U_I}^1 \sim \mathcal{S_I}( [\mathcal{U_I}^0; \texttt{INTER\_I}(\mathcal{U_R}^0)] ). \tag{4}\]

If the output of the inquirer begins or ends with the termination token, the stopping condition, \texttt{stop} is met, and consequently, the algorithm terminates.
Otherwise, the prompt is extracted using \texttt{extract\_prompt}, and the process persists for the coming turns until it reaches the maximum number of turns: $t = max\_t$.
For the $t$-th turn, the utterances will be:
\[\mathcal{U_I}^t \sim \mathcal{S_I} ( \mathcal{T}^0, \mathcal{T}^1,\dots,\mathcal{T}^{t-1} ) \tag{5}\]
\[\mathcal{U_R}^t \sim \mathcal{S_R} ( \mathcal{T}^0, \mathcal{T}^1,\dots,\mathcal{T}^{t-1}, \mathcal{U_I}^t ) \tag{6}\]

\begin{algorithm}
\SetAlgoLined
\SetKwInOut{Input}{Input}
\SetKwInOut{Output}{Output}
\Input{$\mathcal{P}, \mathcal{G}$: persona and goal}
\Output{$\mathcal{D}(\mathcal{S_I}, \mathcal{S_R})$}

$\mathcal{D} \gets \{\};$ \\
$\mathcal{U_I}^0 \gets \mathcal{S_I}(\texttt{SYS\_I}(\mathcal{P}, \mathcal{G}))$ \\
$\mathcal{U_I}^0 := \texttt{extract\_prompt}(\mathcal{U_I}^0)$ \\
\lIf{$\mathcal{U_I}^0 = \emptyset$} {break}
$\mathcal{U_R}^0 \gets \mathcal{S_R}(\mathcal{U_I}^0)$ \\

\For{$t = 1 \rightarrow \text{max\_t}$}{
    $\mathcal{U_I}^t \gets \mathcal{S_I}([\mathcal{U_I}^{t-1}; \texttt{INTER\_I}(\mathcal{U_R}^{t-1})]);$

    \If{\texttt{stop}($\mathcal{U_I}^t$)} {break;}
    
    $\mathcal{U_I}^t := \texttt{extract\_prompt}(\mathcal{U_I}^t)$ \\
    \If{$\mathcal{U_I}^t = \emptyset$} {break;}
    
    $\mathcal{U_R}^t \gets \mathcal{S_R}(\mathcal{D}, \mathcal{U_I}^t);$ \\
    $\mathcal{D} \gets \{\mathcal{D}, [\mathcal{U_I}^t; \mathcal{U_R}^t]\};$ \\
}
\caption{LLM Roleplay}
\label{algo:llm_roleplay}
\end{algorithm}

\begin{table*}
    %\setstretch{1}
    \centering
    \resizebox{0.85\linewidth}{!}{
    %\renewcommand{\arraystretch}{1.8}
    %\makebox[\textwidth]{
    \begin{tabular}{lcccc}
    %\hline
    \toprule
           & \textbf{Llama-2} & \textbf{Mixtral} & \textbf{Vicuna} & \textbf{GPT4} \\
    \midrule
           \textbf{Avg. \# Turns per Dialogue}& 2.62 (1.54) & 3.86 (2.19) & 7.12 (3.79) & \textbf{7.60} (3.08) \\

           \textbf{Avg. \# Tokens per Prompt}& 77.77 (46.20) & \textbf{50.82} (26.47) & 75.19 (92.43) & 68.13 (60.37) \\
           
           \textbf{Avg. \# Tokens per Response}& \textbf{347.50} (142.14) & 302.47 (151.30) & 228.29 (163.92) & 267.69 (147.26) \\
    \midrule
       \textbf{No-prompt}& 6.82\% (1.48\%)& 0.97\% (0.57\%)& 7.90\% (0.54\%)& \textbf{0.17}\% (0.04\%)\\
       
       \textbf{Multiple Prompts}& 8.79\% (0.95\%)& 8.40\% (1.52\%)& \textbf{6.08}\% (0.41\%)& 39.21\% (2.38)\\

       \textbf{Incoherent Response}& 3.12\% (0.30\%)& 0.13\% (0.09\%)& 0.79\% (0.10\%)& \textbf{0.03}\% (0.04\%)\\

       \textbf{Number of Self-Replies}& 5.50\% (3.25\%)& 5.99\% (1.43)& 69.39\% (5.77\%)& \textbf{5.48}\% (0.09\%)\\
       
       \textbf{Incoherent Response (Responder)}& \textbf{0.56}\% (0.57\%)& 1.16\% (0.19\%)& 8.01\% (0.93\%)& 7.50\% (0.35\%)\\       
       \bottomrule
    \end{tabular}}
    %}
    \caption{Analysis of persona-specific dialogue collection (top) and failure cases (bottom) conducted for Llama-2, Mixtral, Vicuna, and GPT4. The results are averaged over runs with three different seeds. We show that GPT4 is successful at holding dialogues with longer utterances and having relatively fewer failure cases. Meanwhile, Mixtral is better at providing short on-point prompts. The number of utterances in dialogues is preferred to be larger, while for other metrics, smaller values are better. The standard deviation is indicated in parentheses.}
    \label{tab:dialog_collection_full_statistics}
\end{table*}

\section{Experiments}

In the first study, we collect dialogues between human inquirers and model responders ($\mathcal{D}(\mathcal{S_I}^h, \mathcal{S_R})$) by engaging participants with various personas ($\mathcal{P}$) in interactions with a chat-tuned LLM intending to achieve a given goal ($\mathcal{G}$).
Utilizing the same set of personas ($\mathcal{P}$) and goals ($\mathcal{G}$) we generate dialogues between model inquirers and the same responder ($\mathcal{D}(\mathcal{S_I}, \mathcal{S_R})$) by employing the LLM Roleplay method (\Cref{subsec:dialog_aggregation}).
We report the statistics of the generated dialogues such as the average number of turns per dialogue, the number of tokens per prompt (inquirer output), and the number of tokens per response (responder output).
Please refer to Appendix \ref{sec_appendix:dialog_collection} for details on the generation parameters.

Furthermore, we investigate the failure cases of the inquirer LLMs, in following the given instruction.
We depict the most common failure cases, report their statistics, and describe their detection mechanisms (\Cref{subsec:failure_cases}).

In the second study, we conduct a human evaluation wherein another set of participants compare the natural $\mathcal{D}(\mathcal{S_I}^h, \mathcal{S_R})$ and the simulated $\mathcal{D}(\mathcal{S_I}, \mathcal{S_R})$ dialogues to discern the simulated counterpart $\mathcal{D}(\mathcal{S_I}, \mathcal{S_R})$ (\Cref{subsec:human_evaluation}).
We report the total and per-model undetectability rates, the utterance number on which the dialogue was detected, and the distribution of the duration and confidence choices.

Moreover, we use generalized linear models to analyze the detection rate of the simulated dialogue, the detection utterances number, and the duration users spent making a choice to analyze how different inquirer LLMs behave.

For our experiments, we used a single NVIDIA A100 GPU with 80GB memory for Llama-2 and Vicuna.
We utilized up to 92\% of the memory.

\subsection{Persona-Specific Dialogue Collection}
\label{subsec:dialog_aggregation}

For this study, participants were instructed to interact with Llama-2\footnote{Due to limited deployment resources quantized Llama-2-13B-chat-GGUF is utilized \cite{touvron2023llama}, employing the Q5\_K\_M quantization method with 5 bits \cite{frantar2023gptq}} ($\mathcal{S_R}$) to accomplish a designated goal.
We additionally ask participants to provide sociodemographic information ($\mathcal{P}$) such as age group, gender, race, level of education, and whether they identify as native English speaker.
We base the choice of sociodemographic features and the options of the features on previous work by \citet{10.5555/3563572.3563588}.
We provide a detailed screenshot of the persona information form interface in \Cref{fig:dialog_aggregation_persona_form}, and a screenshot of our chat interface in \Cref{fig:dialog_aggregation_app_chat} in the \Cref{sec_appendix:dialog_collection}.
In order to cover different conversational goals, we design ten handcrafted multi-hop goals ($\mathcal{G}$) spanning three domains: "Math", "Coding", and "General Knowledge".

We conduct a study involving 20 participants each engaged in tackling 10 goals, resulting in the generation of 200 natural human-chatbot interaction dialogues ($\mathcal{D}(\mathcal{S_I}^h, \mathcal{S_R})$).
We provide the full sociodemographic distribution of the participants in \Cref{fig:app_age_distr} to \Cref{fig:app_english_distr} in  \Cref{sec_appendix:dialog_collection}.

Subsequently, we generate dialogues utilizing our LLM Roleplay method with a set of three state-of-the-art LLMs and one proprietary conversational agent as inquirers ($\mathcal{S_I}$): llama-2-13B-Chat (Llama-2) \cite{touvron2023llama}, Mixtral-8x7B-Instruct-v0.1 (Mixtral) \cite{jiang2024mixtral}, vicuna-13b-v1.5-16k (Vicuna) \cite{peng2023instruction}, and GPT4 \cite{openai2024gpt4}.
See sample natural and generated dialogues using Mixtral inquirer and Llama-2 responder from \Cref{fig:example-dialogue-1} to \Cref{fig:example-dialogue-5} in \Cref{sec_appendix:more_examples}.

In total, the dialogue collection and generation results in the creation of 200 natural human-chatbot ($\mathcal{D}(\mathcal{S_I}^h, \mathcal{S_R})$) and 800 (with 4 inquirers) simulated ($\mathcal{D}(\mathcal{S_I}, \mathcal{S_R})$) dialogue pairs.

We observe that, on average, GPT-4 \cite{openai2024gpt4} is successful at holding longer dialogues with 7.60 turns on average (\Cref{tab:dialog_collection_full_statistics}).
Mixtral \cite{jiang2024mixtral} on the other hand is better at generating shorter and on-point (based on manual analysis) prompts with 50.82 tokens per prompt.
The generated dialogues have an average of 5.30 turns, each consisting of an average of 67.97 tokens per prompt, resulting in longer dialogues than those from the previous work (\Cref{tab:datasets_comparison}).

\subsection{Impact of Persona-Specific Information}
\label{subsec:persona_ablation_study}
We assess how adding sociodemographic information to the persona description that is provided to the model affects the lexical diversity of the simulated user utterances.
We compare three scenarios: (a) using a baseline prompt with no sociodemographic information in the persona description, (b) adding single sociodemographic features, (c) and including all features as described in our method.
\Cref{tab:lex_diversity_scores} shows the respective Type-Token Ratio (TTR) \cite{zipf2013psycho} and Distinct-N (dist-n) \cite{li-etal-2016-diversity} values quantifying lexical diversity for Mixtral inquirer.
We observe that adding single sociodemographic features as well as combined sociodemographic features to the model prompt significantly increases lexical diversity.
% and that the highest lexical diversity is achieved using all features in combination.
% We report additional measures of lexical diversity in \Cref{tab:app_lex_diversity_scores} in \Cref{sec_appendix:dialog_collection}.

\begin{table}
    \centering
    \resizebox{0.8\columnwidth}{!}{
    \begin{tabular}{lcll}
    \toprule
         \textbf{SD Information}& \textbf{TTR}& \textbf{dist-1}&\textbf{dist-2}\\
    \midrule
         None&   0.281  (0.004) & 0.284 (0.003)& 0.625 (0.007)\\
    \midrule
         Age&   0.572 (0.010) & 0.576 (0.010)& \textbf{0.892} (0.007)\\
         Race&   0.582 (0.006) & 0.587 (0.005)& 0.880 (0.002)\\
         Gender&   0.411 (0.003) & 0.415 (0.003)& 0.747 (0.002)\\
         Education&   0.579 (0.015) & 0.578 (0.016)& 0.869 (0.005)\\
         Is Native EN Speaker&   0.394 (0.003) & 0.394 (0.003)& 0.721 (0.005)\\
    \midrule
         All&  \textbf{0.606} (0.019) & \textbf{0.605} (0.019)& 0.872 (0.006)\\
    \end{tabular}}
    \caption{Ablation study on the impact of sociodemographic (SD) information on the lexical diversity of the simulated user's language measured via mean Type-Token Ratio (TTR) and Distinct-N (dist-n) along with corresponding variance reported in parentheses.
    We observe enhanced lexical diversity with the addition of individual sociodemographic features.}
    % and maximal diversity when combining all features.}
    \label{tab:lex_diversity_scores}
\end{table}

\subsection{Failure Cases}
\label{subsec:failure_cases}
Since the outputs of the LLMs are free-form, applying deterministic functions to their output proves challenging, resulting in a theoretically infinite number of turns.
Therefore, there is a need for a set limit on the number of turns: $max\_t$.
Nevertheless, we try to capture the algorithm failure cases to have an automatic assessment of the inquirer LLM failure cases.
We expect the inquirer LLM to provide the intended prompt enclosed in double quotes as we explicitly request this in our instructions.
See a sample expected output and the failure cases examples in \Cref{table:failure_cases_examples} in \Cref{sec_appendix:failure_cases}.
All of the LLMs in our experiments face the following issues.
\paragraph{Prompt not in double-quotes.} The model fails to provide a prompt within double quotes, thus causing the dialogue to terminate.

\paragraph{Incoherent output.} The responder produces a repetitive token sequence.
We identify such outputs and preemptively end the dialogue.
We utilize the \texttt{incoherent} function to analyze text for incoherent strings (see \Cref{algo:incoherent} in \Cref{sec_appendix:failure_cases}).

\paragraph{Incoherent output of responder.} The inquirer model fails to detect the case when the responder outputs incoherent text, leading to unsuccessful dialogues.
We spot such outputs and stop the dialog, using the same \texttt{incoherent} function.

\paragraph{Inquirer self-reply.} The inquirer model fails to maintain its intended role and answers its own question, resulting in a fully generated dialogue in a single utterance.
We detect this deterministically by examining the presence of any special tokens of the responder model within the output.
For instance, in the case of Llama-2, this token appears as "[INST", while for Vicuna, it manifests as "\#\#\# Human:".

\paragraph{Multiple prompts.} The inquirer outputs multiple strings enclosed in double quotes.
As sometimes this overlaps with the previous case (inquirer self-reply), we select the first as the prompt.

\paragraph{Dialogue-stopping criterion failure.} In the intermediate prompts, we ask the chatbot to output a pre-defined token when it "thinks" the goal assigned to it is achieved.
Nonetheless, it follows a limited set of tokens; for instance, \texttt{"FINISH"} was utilized in our experiments.

\begin{table*}
    \centering
    \resizebox{0.9\linewidth}{!}{
    \begin{tabular}{llcccc}
    \toprule
        \textbf{Subset} &  & \textbf{Llama-2} & \textbf{Mixtral} & \textbf{Vicuna} & \textbf{GPT4} \\
    \midrule
        \multirow{4}{*}{\rotatebox[origin=c]{90}{\textit{total}}} & \textbf{Undetectability Rate}& 33.5\% & \textbf{44.0}\% & 22.5\% & 35.0\% \\
        & \textbf{Confidence: "very confident"}& 33  (16.50\%) & 32  (16.00\%) & 75  (37.50\%) & 43  (21.50\%) \\
        & \textbf{Confidence: "confident"}& 108  (54.00\%) & 83  (41.50\%) & 75  (37.50\%) & 97  (48.50\%) \\
        & \textbf{Confidence: "somewhat confident"}& 59  (42.50\%) & 85  (25.00\%) & 50  (20.00\%) & 60  (30.00\%) \\
    \midrule
        \multirow{5}{*}{\rotatebox[origin=c]{90}{\textit{detected}}} & \textbf{Duration}& \textbf{86.10 (157.14)} & 99.45 (98.81) & 94.63 (150.70) & 124.88 (227.36) \\
        & \textbf{Utterance Number}& 1.72 (0.78) & 2.38 (1.35) & 2.50 (1.84) & \textbf{3.72 (2.35)} \\
        & \textbf{Confidence: "very confident"}& 24  (12.00\%) & 19 (\textbf{9.50\%}) & 65  (32.50\%) & 32  (16.00\%) \\
        & \textbf{Confidence: "confident"}& 80  (40.00\%) & 51 (\textbf{25.50\%}) & 59  (29.50\%) & 66  (33.00\%) \\
        & \textbf{Confidence: "somewhat confident"}& 29  (14.50\%) & 42  (\textbf{21.00}\%) & 31  (15.50\%) & 32  (16.00\%) \\
    \midrule
        \multirow{4}{*}{\rotatebox[origin=c]{90}{\textit{undetected}}} & \textbf{Duration}& \textbf{120.44 (139.93)} & 114.93 (157.46) & 62.24 (62.37) & 94.52 (124.03) \\
        & \textbf{Confidence: "very confident"}& 9  (4.50\%) & 13 (\textbf{6.50}\%) & 10 (5.00\%) & 11  (5.50\%) \\
        & \textbf{Confidence: "confident"}& 28 (14.00\%) & 32 (\textbf{16.00}\%) & 16 (8.00\%) & 31  (15.50\%)  \\
        & \textbf{Confidence: "somewhat confident"}& 30 (15.00\%) & 43 (\textbf{21.50}\%) & 19 (9.50\%) & 28 (14.00\%) \\
   \bottomrule
    \end{tabular}}
    \caption{Analysis of human-evaluation results for detected, undetected, and total dialogues for Llama-2, Mixtral, Vicuna, and GPT4, showing confidence statistics as occurrences (percentages in parentheses).
    We exhibit that Mixtral has the highest undetectability rate of 44\% (\textbf{50\%} reflecting absolute indistinguishability). Moreover, it has the lowest confidence choice statistics for not confidently detected and the highest statistics for confidently undetected dialogues.
    GPT4 has the highest utterance number: 3.72, showing that it is identified in later utterances.
    The duration (in seconds) and the utterance number standard deviations are in parentheses.}
    \label{tab:human_evaluation_full_statistics}
\end{table*}

We attribute these failures to two common issues in LLMs: limited context length and a restricted set of fine-tuned instructions \cite{kaddour2023challenges}.

\paragraph{Toxic content detection.}
To proactively address the generation of potentially harmful content, we employ the Llama-2 Guard model \cite{metallamaguard2} to filter out toxic dialogues.

In most cases, GPT4 outperforms other models with lower failure rates: 0.17\% for responses without prompts, 0.03\% for incoherent responses, and 5.48\% for self-replies (\Cref{tab:dialog_collection_full_statistics}).
However, it struggles with providing a single prompt, failing 39.21\% of the time.
In contrast, Vicuna excels in generating single prompts, achieving a failure rate of 6.08\%.
Llama-2 receives fewer incoherent outputs, with 0.56\%.
Mixtral’s performance is intermediate, showing more balanced results across different metrics.
Additional plots and detailed data are provided in \Cref{sec_appendix:dialog_collection}.
In these experiments, we detected no unsafe content.

\subsection{Human-Evaluation}
\label{subsec:human_evaluation}

To answer the question of how well LLM inquirer and responder dialogues $\mathcal{D}(\mathcal{S_I}, \mathcal{S_R})$ approximate dialogues between human inquirer and responder $\mathcal{D}(\mathcal{S_I}^h, \mathcal{S_R})$, we conduct a human evaluation study.
In each round, participants are shown two dialogues side by side, both featuring the same persona ($\mathcal{P}$) and solving the same goal ($\mathcal{G}$).

We allocated different sets of dialogue pairs for each participant, ensuring that no participant encounters multiple dialogues from the same user (from \Cref{subsec:dialog_aggregation}) solving the same goal.
A new group of 20 participants was selected, with each participant tasked with reviewing 40 dialogue pairs.
The participants selected for this study represent a wide range of occupational backgrounds.
Some have no prior experience with chatbots, while others are industry professionals.

Participants are required to answer three questions for each dialogue pair:
(i) Identify which dialogues they perceive as artificial (simulated): Choices include "1st (left)", "2nd (right)", or "Not sure" the latter is considered to be a tie.
(ii) Express the level of confidence in their selection: Options are "Somewhat Confident," "Confident," and "Very Confident".
(iii) Identify the specific utterance number that signifies the artificiality within the dialogue: Options depend on the number of utterances of the dialogue pairs.
We provide a detailed view of the instructions provided to users in \Cref{fig:human_evaluation_start} and the interfaces of the applications used for the study \Cref{fig:human_evaluation_choice} in \Cref{sec_appendix:human_evaluation}.

In conducting the human evaluation, we also track the amount of time participants take to respond to questions.
This measure allows us to estimate a proxy measure of the complexity of the dialogue pairs.
We assume that the longer it takes for a participant to make a decision, the more challenging the pair is, indicating that the simulated dialogue is difficult to discern.

\begin{figure*}
    \centering
\includegraphics[width=0.65\linewidth]{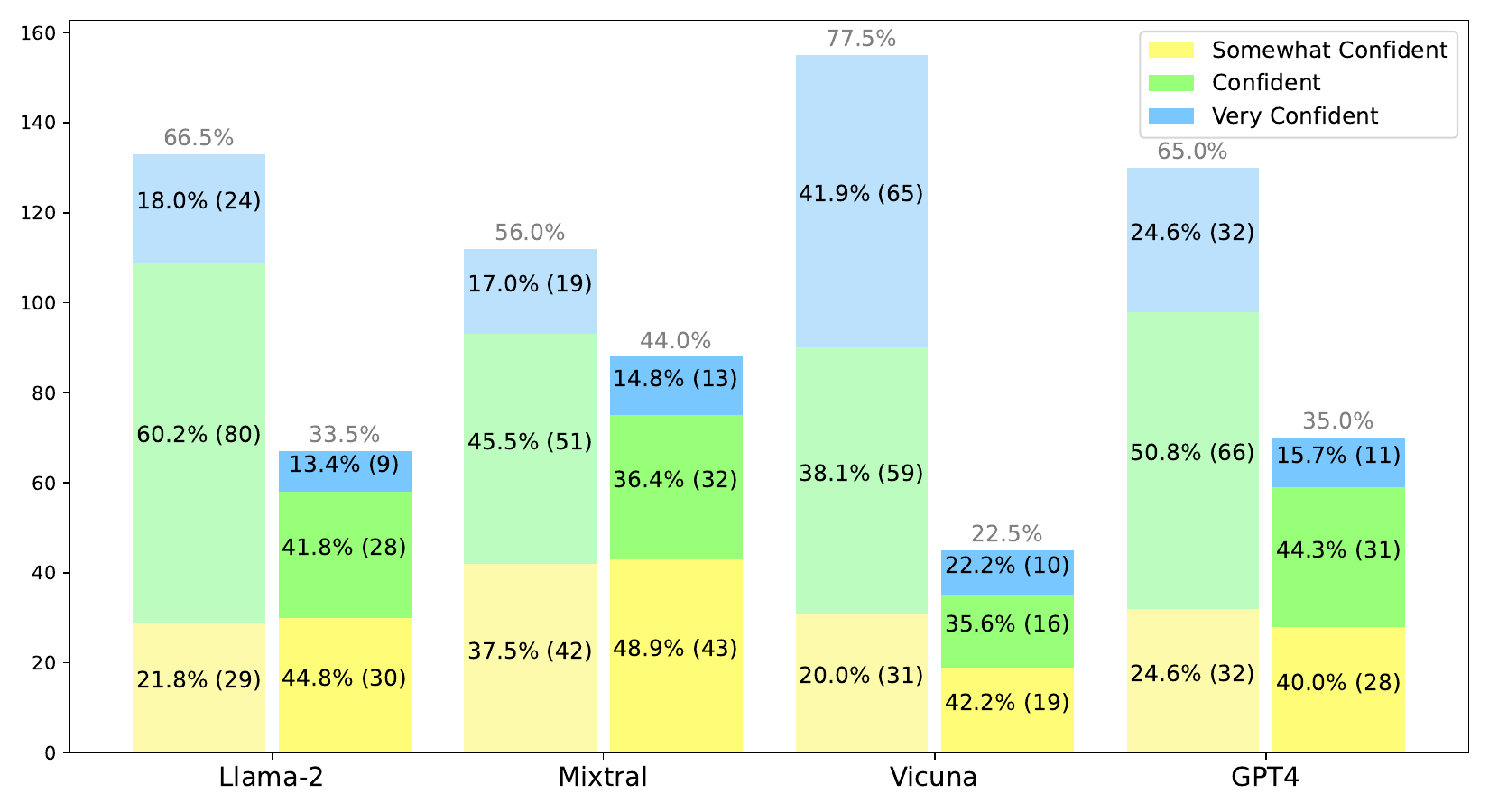}
    \caption{The distribution of detectability (left) and undetectability rates (right) per model for Llama-2, Mixtral, Vicuna, and GPT4.
    Each bar is stacked with confidence levels of: "Somewhat confident", "Confident" and "Very Confident".
    We shot that Mixtral has a relatively high undetectability rate of 44\%, followed by GPT4 at 35\%, Llama-2 at 33.5\%, and Vicuna at 22.5\%.
    The total (un)detectability %and undetectability 
    rates for each model are mentioned in gray.}
    \label{fig:human_evaluation_choices_per_model}
\end{figure*}

The study shows that among the 800 samples, the simulated dialogues remained undetected in 33.75\% (\textbf{50\%} reflecting absolute indistinguishability) of the dialogues.
Per model, statistics show that Mixtral has the highest undetectability rate of 44.0\%, after which is GPT4 with 35.0\% followed by Llama-2 and Vicuna with 33.5\% and 22.5\% respectively (\Cref{fig:human_evaluation_choices_per_model}).
See the per-confidence choice distribution in \Cref{fig:human_evaluation_choices} in \Cref{sec_appendix:human_evaluation}.

The highest utterance number on the detected set of dialogues has GPT4, meaning that it took more utterances for participants to recognize the simulated nature of the inquirer's responses (\Cref{tab:human_evaluation_full_statistics}).
For the detected subset of dialogues, Mixtral has the highest percentages for all confidence choices: 9.50\%, 25.50\%, and 21.00\% for "very confident", "confident" and "somewhat confident".
Also, it is the best for the undetected pair of dialogues with 6.50\%, 16.00\%, and 21.50\% respectively.
Llama-2 excels in duration with 86.10 and 120.44 seconds for the detected and undetected dialogue pairs.
However, it should be noted that the variation in duration is high, indicating that the participants did not complete the study at a consistent pace.

To further investigate how various instruction-tuned LLMs behave as inquirers within this setup we analyze the simulated dialogue detection probability, the detection utterance number, and the duration participants spent, using generalized linear mixed models (GLMMs), with the choice of LLM as the independent variable.
We additionally include random effects to account for potential confounding effects of individual participants' detection abilities and users from \Cref{subsec:dialog_aggregation}.

We find significant effects of the choice of the model on the detection probability and the utterance position at which the participants formed their decision.
We summarize the results of our statistical analyses in \Cref{tab:human_evaluation_feature_significance} in \Cref{sec_appendix:human_evaluation} using CLD codings \citep{piepho2004algorithm} and discuss our key findings in the following.

\paragraph{Effects on Detection Probability.}
We fit a binomial model (logit link) to predict the detection probability depending on the model.
Concretely, we estimate a GLMM specified by: $detection\_rate \sim model + (1| participant) + (1| generator\_user)$.
We observe a significant effect of the choice of the model on the detection probability ($\chi^2(3)$=21.49, $p < 0.001$).
A post hoc Wald comparison of the contrasts for model types revealed significant differences between Vicuna and all other models (Vicuna being most likely to be detected), and Mixtral and LLama-2 (Mixtral being significantly less likely to be detected).
We do not find a significant difference between GPT4 and Mixtral.

\paragraph{Effects on Utterance Positions.}
For the position of decision-forming utterances, we fit a respective GLMM and find a significant effect of the choice of the model ($F$=31.73, $p < 0.001$).
A post hoc Wald comparison of the contrasts for model types revealed significantly higher utterance positions for GPT4 than for the other models and that Llama-2 received significantly lower utterance position ratings.
We do not observe a significant difference between Mixtral and Vicuna.

% Overall, our findings suggest that Mixtral, an open-source LLM, demonstrates better performance in embodying a specified persona and simulating human-chatbot interactions compared to GPT4 in terms of detection probabilities.
% Nonetheless, when considering the detection of simulated dialogues based on the detection utterance number, GPT4 shows better performance which is likely to be attributable to GPT4 generating dialogues with more utterances.

Our findings suggest that Mixtral, an open-source LLM, performs better than GPT4 in embodying a specified persona and simulating human-chatbot interactions in terms of detection probabilities.
However, in detecting simulated dialogues based on utterance number, GPT4 outperforms, likely due to generating more utterances.

\section{Related Work}

\begin{table*}
    \centering
    \resizebox{1.0\linewidth}{!}{
    \begin{tabular}{l|lcccccc}
    \toprule
        \textbf{Type} & \textbf{Dataset Name} & \textbf{\# Dialogues} & \textbf{Avg. \# Turns/Dialogue} & \textbf{Avg. \# Tokens/Prompt} & \textbf{Avg. \# Tokens/Response} & \textbf{Topics} & \textbf{Personalized} \\ 
    \midrule
        \multirow{3}{*}{\rotatebox[origin=c]{90}{\footnotesize natural}} & \textbf{OpenAssistant} \cite{kopf2023openassistant} & 3k & 2.12 & 28.28 & 171.34 & human-crafted & \textbf{yes} \\
    \cmidrule(lr){2-8}
        & \textbf{LMSYS-Chat-1M} \cite{zheng2024lmsyschatm} & 777k & 1.92 & 55.23 & 163.66 & human-crafted & \textbf{yes} \\
        & WildChat \cite{zhao2024inthewildchat} & 360k & 2.46 & 164.32 & 276.50 & \textbf{open} & no \\
    \midrule
        \multirow{2}{*}{\rotatebox[origin=c]{90}{\footnotesize synthetic}}
        & \textbf{UltraChat} \cite{ding-etal-2023-enhancing} & 1.5M & 3.85 & 52.54 & 249.41 & model-generated & no \\
    \cmidrule(lr){2-8}
        & \textbf{LLM Roleplay (Ours)} & any & 5.30 (2.11) & 67.97 (10.51) & 286.48 (151.15) & \textbf{any} & \textbf{yes} \\

   \bottomrule
    \end{tabular}}
    \caption{Statistics of key human-crafted (natural) and model-generated (synthetic) datasets (English subsets) relevant to our study. Our method can generate unlimited dialogues across various topics, featuring persona-based prompts and longer utterances. The natural datasets include personalized information reflecting the inquirer’s subjectivity. Standard deviations are shown in parentheses. For a comprehensive dataset list, see \Cref{app_tab:datasets_comparison_extensive} in \Cref{sec_appendix:more_related_work}.}
    \label{tab:datasets_comparison}
\end{table*}

\paragraph{Conversational Datasets.}
Approximation of the true distribution of human-chatbot dialogues is challenging and a significant number of diverse human annotators are needed.
Incorporating participants from different sociodemographic profiles addressing a given conversational goal can substantially enhance the richness of the dataset.
However, the associated time and resource requirements render the respectively needed large-scale studies infeasible for many researchers.
Alternatively, a less resource-intensive and faster method for collecting human-chatbot conversations is to generate them using LLMs \cite{zhang2024instruction}.
This method has been utilized in various setups for dialogue generation, such as human-human \cite{adiwardana2020humanlike, kim-etal-2023-soda, chen-etal-2023-places, li2023camel}, teacher-student \cite{macina-etal-2023-mathdial}, and patient-physician \cite{wang2023notechat}.

Prior work proposed a large number of human-crafted and synthetically generated datasets, each trying to collect more dialogue pairs revolving around various topics (\Cref{tab:datasets_comparison}).
OpenAssistant \cite{kopf2023openassistant} collects human-crafted conversational dialogue trees, with prompts and answers generated by different humans.
LMSYS-Chat-1M \cite{zheng2024lmsyschatm} contains the refined version of dialogue logs collected via an online chat interface and predominantly contains dialogues between users and Vicuna-13b \cite{peng2023instruction}.
UltraChat \cite{ding-etal-2023-enhancing} and GLAN \cite{li2024synthetic} synthetically generate dialogues using proprietary conversational AI systems (ChatGPT Turbo) evolving around topics generated by the same systems.

\paragraph{Personas in Large Language Models.}
Large language models (LLMs) have demonstrated distinct behaviors and personas \cite{andreas-2022-language, wolf2024fundamental}.
\citet{andreas-2022-language} note that LLMs interpret behaviors from text prompts, influencing generated content.
Additionally, \citet{wolf2024fundamental} argue that LLMs function as mixture decompositions, where prompts shift component balances, thus triggering persona-specific responses.
Furthermore, \citet{beck-etal-2024-sensitivity} investigate the effects of sociodemographic prompts on model responses, finding that while beneficial in some settings, they produce varied effects across models.

Building on these findings, our work introduces a novel, goal-oriented, persona-centric method for generating diverse, multi-turn dialogues.
This method simulates dialogues across various combinations of conversational goals, personas, and LLMs, enabling the creation of countless simulated dialogues without theoretical limits.

\section{Discussion and Future Work}
In this work, we propose the novel LLM Roleplay method: an automatic, model-agnostic approach for eliciting multi-turn, goal-oriented, persona-based simulated human-chatbot dialogues.
We develop and validate our method through two user studies involving 40 participants.
Our findings show that up to 44\% of these dialogues, where 50\% represents perfect indistinguishability, are indistinguishable from real human-chatbot interactions and feature more turns than previous datasets.

Building on this work’s findings, several future research avenues warrant exploration.
First, our method can generate user-specific conversational datasets for targeted alignment and domain-adaptation (e.g., using RLHF \cite{christiano2023deep}).
Second, it can improve dialogue evaluation by providing ample realistic conversational data.

While this paper presents important findings on LLM Roleplay and provides the first evidence that our method can approximate human-chatbot dialogues, the sociodemographic representativeness of our method still needs assessment and advancement.
We release our dialogue dataset, method code, and synthesized dialogues to support future research in this promising field.

\section{Conclusion}
We present our novel LLM Roleplay method for simulating human-chatbot interaction.
In a series of two user studies, we collect real human-chatbot dialogues and demonstrate that LLM Roleplay can generate diverse multi-turn conversations that approximate natural human-chatbot dialogues with a high level of indistinguishability.
Our findings highlight the potential of LLMs in simulating human-chatbot interactions to synthesize realistic dialogues that create new opportunities for real-time model evaluation and training data generation for model fine-tuning.

\FloatBarrier

\section{Acknowledgments}
This work has been funded by the German Research Foundation (DFG) as part of the UKP-SQuARE project (grant GU 798/29-1).
This work has been funded by the LOEWE Distinguished Chair “Ubiquitous Knowledge Processing”, LOEWE initiative, Hesse, Germany (Grant Number: LOEWE/4a//519/05/00.002(0002)/81).
We gratefully acknowledge the support of Microsoft with a grant for access to OpenAI GPT models via the Azure cloud (Accelerate Foundation Model Academic Research).

\section*{Limitations}
In this section, we explore the inherent limitations of this research study.
We further note that all our experiments have been approved by the local ethics reviewing board at Technical University Darmstadt.

\paragraph{Sociodemographic Representativeness.}
An important limitation of our study lies in the sociodemographic distribution of the participants involved in the dialogue collection process.
The pool of our studies' participants serves as an initial investigation into the promising direction of interaction simulation and cannot represent the full spectrum of sociodemographic backgrounds.
To address this limitation, future research needs to broaden the scope of participants' sociodemographic groups and assess the replicability of our findings within large user studies, encompassing a more comprehensive range of sociodemographic groups.
Studies should be conducted with specific sociodemographic groups, ensuring that participants representing a broader set of combinations of the persona features described in the paper are covered (e.g. via crowdsourcing on platforms like Prolific).
By doing so, a more nuanced understanding of natural dialogue dynamics across diverse populations can be achieved, and allow us to uncover weaknesses and respectively needed improvements building upon our initial method specification.

\paragraph{Controllability.}
In this work, we have endeavored to detect and prevent failure cases of the proposed method.
Despite our efforts, it is important to acknowledge that achieving absolute coverage in detecting all potential failure cases remains elusive.
For example, instances where multiple turns of appreciation occur bilaterally present a challenge that we have not fully addressed.
Moreover, we strive to save all detected failure cases by re-generating with different parameters.
However, this approach has not proven to be effective.
Future research should concentrate on exploring these scenarios further to detect and prevent failure cases more efficiently.

Additionally, although our LLM Roleplay shows promising results in our experimental settings, it is not immune to the common challenges associated with large language models.
Problems like hallucinations and associated LLM behavior issues can still arise, even though we have not encountered any during our experiments.

\section*{Ethics Statement}
As discussed in the previous section, the narrow sociodemographic spectrum of participants involved in our user studies demands follow-up work to study the generated dialogues' sociodemographic validity.
Further, the sociodemographic information provided in our method's prompts could potentially trigger biased content generation within the underlying LLM.
As a first countermeasure, our method comprises a guard model to prevent the generation of toxic content.
We, however, note that biases can also manifest more subtly and want to emphasize that, while we did not observe any such cases, future work should carefully assess whether such effects are present in future LLMs.

While synthetic data generation always entails a risk of generating invalid or biased data, we argue that our work takes an important step toward more valid user data generation. 
All our user studies have been approved by the local ethics reviewing board.

\bibliography{anthology,custom}

\begin{thebibliography}{46}
\providecommand{\natexlab}[1]{#1}

\bibitem[{Adiwardana et~al.(2020)Adiwardana, Luong, So, Hall, Fiedel, Thoppilan, Yang, Kulshreshtha, Nemade, Lu, and Le}]{adiwardana2020humanlike}
Daniel Adiwardana, Minh-Thang Luong, David~R. So, Jamie Hall, Noah Fiedel, Romal Thoppilan, Zi~Yang, Apoorv Kulshreshtha, Gaurav Nemade, Yifeng Lu, and Quoc~V. Le. 2020.
\newblock \href {https://arxiv.org/abs/2001.09977} {Towards a human-like open-domain chatbot}.
\newblock \emph{Preprint}, arXiv:2001.09977.

\bibitem[{Almazrouei et~al.(2023)Almazrouei, Alobeidli, Alshamsi, Cappelli, Cojocaru, Debbah, Étienne Goffinet, Hesslow, Launay, Malartic, Mazzotta, Noune, Pannier, and Penedo}]{almazrouei2023falcon}
Ebtesam Almazrouei, Hamza Alobeidli, Abdulaziz Alshamsi, Alessandro Cappelli, Ruxandra Cojocaru, Mérouane Debbah, Étienne Goffinet, Daniel Hesslow, Julien Launay, Quentin Malartic, Daniele Mazzotta, Badreddine Noune, Baptiste Pannier, and Guilherme Penedo. 2023.
\newblock \href {https://arxiv.org/abs/2311.16867} {The falcon series of open language models}.
\newblock \emph{Preprint}, arXiv:2311.16867.

\bibitem[{Andreas(2022)}]{andreas-2022-language}
Jacob Andreas. 2022.
\newblock \href {https://doi.org/10.18653/v1/2022.findings-emnlp.423} {Language models as agent models}.
\newblock In \emph{Findings of the Association for Computational Linguistics: EMNLP 2022}, pages 5769--5779, Abu Dhabi, United Arab Emirates. Association for Computational Linguistics.

\bibitem[{Beck et~al.(2024)Beck, Schuff, Lauscher, and Gurevych}]{beck-etal-2024-sensitivity}
Tilman Beck, Hendrik Schuff, Anne Lauscher, and Iryna Gurevych. 2024.
\newblock \href {https://aclanthology.org/2024.eacl-long.159} {Sensitivity, performance, robustness: Deconstructing the effect of sociodemographic prompting}.
\newblock In \emph{Proceedings of the 18th Conference of the European Chapter of the Association for Computational Linguistics (Volume 1: Long Papers)}, pages 2589--2615, St. Julian{'}s, Malta. Association for Computational Linguistics.

\bibitem[{Chen et~al.(2023)Chen, Papangelis, Tao, Kim, Rosenbaum, Liu, Yu, and Hakkani-Tur}]{chen-etal-2023-places}
Maximillian Chen, Alexandros Papangelis, Chenyang Tao, Seokhwan Kim, Andy Rosenbaum, Yang Liu, Zhou Yu, and Dilek Hakkani-Tur. 2023.
\newblock \href {https://doi.org/10.18653/v1/2023.findings-eacl.63} {{PLACES}: Prompting language models for social conversation synthesis}.
\newblock In \emph{Findings of the Association for Computational Linguistics: EACL 2023}, pages 844--868, Dubrovnik, Croatia. Association for Computational Linguistics.

\bibitem[{Christiano et~al.(2017)Christiano, Leike, Brown, Martic, Legg, and Amodei}]{christiano2023deep}
Paul~F. Christiano, Jan Leike, Tom~B. Brown, Miljan Martic, Shane Legg, and Dario Amodei. 2017.
\newblock \href {https://proceedings.neurips.cc/paper/2017/hash/d5e2c0adad503c91f91df240d0cd4e49-Abstract.html} {Deep reinforcement learning from human preferences}.
\newblock In \emph{Advances in Neural Information Processing Systems 30: Annual Conference on Neural Information Processing Systems 2017, December 4-9, 2017, Long Beach, CA, {USA}}, pages 4299--4307.

\bibitem[{Ding et~al.(2023)Ding, Chen, Xu, Qin, Hu, Liu, Sun, and Zhou}]{ding-etal-2023-enhancing}
Ning Ding, Yulin Chen, Bokai Xu, Yujia Qin, Shengding Hu, Zhiyuan Liu, Maosong Sun, and Bowen Zhou. 2023.
\newblock \href {https://doi.org/10.18653/v1/2023.emnlp-main.183} {Enhancing chat language models by scaling high-quality instructional conversations}.
\newblock In \emph{Proceedings of the 2023 Conference on Empirical Methods in Natural Language Processing}, pages 3029--3051, Singapore. Association for Computational Linguistics.

\bibitem[{Frantar et~al.(2023)Frantar, Ashkboos, Hoefler, and Alistarh}]{frantar2023gptq}
Elias Frantar, Saleh Ashkboos, Torsten Hoefler, and Dan Alistarh. 2023.
\newblock \href {https://arxiv.org/abs/2210.17323} {Gptq: Accurate post-training quantization for generative pre-trained transformers}.
\newblock \emph{Preprint}, arXiv:2210.17323.

\bibitem[{Gopalakrishnan et~al.(2019)Gopalakrishnan, Hedayatnia, Chen, Gottardi, Kwatra, Venkatesh, Gabriel, and Hakkani{-}T{\"{u}}r}]{gopalakrishnan2023topicalchat}
Karthik Gopalakrishnan, Behnam Hedayatnia, Qinglang Chen, Anna Gottardi, Sanjeev Kwatra, Anu Venkatesh, Raefer Gabriel, and Dilek Hakkani{-}T{\"{u}}r. 2019.
\newblock \href {https://doi.org/10.21437/Interspeech.2019-3079} {Topical-chat: Towards knowledge-grounded open-domain conversations}.
\newblock In \emph{Interspeech 2019, 20th Annual Conference of the International Speech Communication Association, Graz, Austria, 15-19 September 2019}, pages 1891--1895. {ISCA}.

\bibitem[{Gunasekar et~al.(2024)Gunasekar, Zhang, Aneja, Mendes, Giorno, Gopi, Javaheripi, Kauffmann, de~Rosa, Saarikivi, Salim, Shah, Behl, Wang, Bubeck, Eldan, Kalai, Lee, and Li}]{gunasekar2024textbooks}
Suriya Gunasekar, Yi~Zhang, Jyoti Aneja, Caio Cesar~Teodoro Mendes, Allie~Del Giorno, Sivakanth Gopi, Mojan Javaheripi, Piero~Conti Kauffmann, Gustavo~Henrique de~Rosa, Olli Saarikivi, Adil Salim, Shital Shah, Harkirat Behl, Xin Wang, Sebastien Bubeck, Ronen Eldan, Adam~Tauman Kalai, Yin~Tat Lee, and Yuanzhi Li. 2024.
\newblock \href {https://openreview.net/forum?id=Fq8tKtjACC} {Textbooks are all you need}.

\bibitem[{Jiang et~al.(2024)Jiang, Sablayrolles, Roux, Mensch, Savary, Bamford, Chaplot, de~las Casas, Hanna, Bressand, Lengyel, Bour, Lample, Lavaud, Saulnier, Lachaux, Stock, Subramanian, Yang, Antoniak, Scao, Gervet, Lavril, Wang, Lacroix, and Sayed}]{jiang2024mixtral}
Albert~Q. Jiang, Alexandre Sablayrolles, Antoine Roux, Arthur Mensch, Blanche Savary, Chris Bamford, Devendra~Singh Chaplot, Diego de~las Casas, Emma~Bou Hanna, Florian Bressand, Gianna Lengyel, Guillaume Bour, Guillaume Lample, Lélio~Renard Lavaud, Lucile Saulnier, Marie-Anne Lachaux, Pierre Stock, Sandeep Subramanian, Sophia Yang, Szymon Antoniak, Teven~Le Scao, Théophile Gervet, Thibaut Lavril, Thomas Wang, Timothée Lacroix, and William~El Sayed. 2024.
\newblock \href {https://arxiv.org/abs/2401.04088} {Mixtral of experts}.
\newblock \emph{Preprint}, arXiv:2401.04088.

\bibitem[{Kaddour et~al.(2023)Kaddour, Harris, Mozes, Bradley, Raileanu, and McHardy}]{kaddour2023challenges}
Jean Kaddour, Joshua Harris, Maximilian Mozes, Herbie Bradley, Roberta Raileanu, and Robert McHardy. 2023.
\newblock \href {https://arxiv.org/abs/2307.10169} {Challenges and applications of large language models}.
\newblock \emph{Preprint}, arXiv:2307.10169.

\bibitem[{Kim et~al.(2023)Kim, Hessel, Jiang, West, Lu, Yu, Zhou, Bras, Alikhani, Kim, Sap, and Choi}]{kim-etal-2023-soda}
Hyunwoo Kim, Jack Hessel, Liwei Jiang, Peter West, Ximing Lu, Youngjae Yu, Pei Zhou, Ronan Bras, Malihe Alikhani, Gunhee Kim, Maarten Sap, and Yejin Choi. 2023.
\newblock \href {https://doi.org/10.18653/v1/2023.emnlp-main.799} {{SODA}: Million-scale dialogue distillation with social commonsense contextualization}.
\newblock In \emph{Proceedings of the 2023 Conference on Empirical Methods in Natural Language Processing}, pages 12930--12949, Singapore. Association for Computational Linguistics.

\bibitem[{K{\"o}pf et~al.(2023)K{\"o}pf, Kilcher, von R{\"u}tte, Anagnostidis, Tam, Stevens, Barhoum, Nguyen, Stanley, Nagyfi, ES, Suri, Glushkov, Dantuluri, Maguire, Schuhmann, Nguyen, and Mattick}]{kopf2023openassistant}
Andreas K{\"o}pf, Yannic Kilcher, Dimitri von R{\"u}tte, Sotiris Anagnostidis, Zhi~Rui Tam, Keith Stevens, Abdullah Barhoum, Duc~Minh Nguyen, Oliver Stanley, Rich{\'a}rd Nagyfi, Shahul ES, Sameer Suri, David~Alexandrovich Glushkov, Arnav~Varma Dantuluri, Andrew Maguire, Christoph Schuhmann, Huu Nguyen, and Alexander~Julian Mattick. 2023.
\newblock \href {https://openreview.net/forum?id=VSJotgbPHF} {Openassistant conversations - democratizing large language model alignment}.
\newblock In \emph{Thirty-seventh Conference on Neural Information Processing Systems Datasets and Benchmarks Track}.

\bibitem[{Kumar et~al.(2021)Kumar, Kelley, Consolvo, Mason, Bursztein, Durumeric, Thomas, and Bailey}]{10.5555/3563572.3563588}
Deepak Kumar, Patrick~Gage Kelley, Sunny Consolvo, Joshua Mason, Elie Bursztein, Zakir Durumeric, Kurt Thomas, and Michael Bailey. 2021.
\newblock Designing toxic content classification for a diversity of perspectives.
\newblock In \emph{Proceedings of the Seventeenth USENIX Conference on Usable Privacy and Security}, SOUPS'21, USA. USENIX Association.

\bibitem[{Li et~al.(2023{\natexlab{a}})Li, Hammoud, Itani, Khizbullin, and Ghanem}]{li2023camel}
Guohao Li, Hasan Abed Al~Kader Hammoud, Hani Itani, Dmitrii Khizbullin, and Bernard Ghanem. 2023{\natexlab{a}}.
\newblock \href {https://openreview.net/forum?id=3IyL2XWDkG} {{CAMEL}: Communicative agents for ''mind'' exploration of large language model society}.
\newblock In \emph{Thirty-seventh Conference on Neural Information Processing Systems}.

\bibitem[{Li et~al.(2024)Li, Dong, Tang, Wang, Zhang, Huang, Huang, Huang, Huang, Zhang, Gu, Cheng, Wang, Chen, Dong, Lu, Sui, Wang, Lam, and Wei}]{li2024synthetic}
Haoran Li, Qingxiu Dong, Zhengyang Tang, Chaojun Wang, Xingxing Zhang, Haoyang Huang, Shaohan Huang, Xiaolong Huang, Zeqiang Huang, Dongdong Zhang, Yuxian Gu, Xin Cheng, Xun Wang, Si-Qing Chen, Li~Dong, Wei Lu, Zhifang Sui, Benyou Wang, Wai Lam, and Furu Wei. 2024.
\newblock \href {https://arxiv.org/abs/2402.13064} {Synthetic data (almost) from scratch: Generalized instruction tuning for language models}.
\newblock \emph{ArXiv preprint}, abs/2402.13064.

\bibitem[{Li et~al.(2016)Li, Galley, Brockett, Gao, and Dolan}]{li-etal-2016-diversity}
Jiwei Li, Michel Galley, Chris Brockett, Jianfeng Gao, and Bill Dolan. 2016.
\newblock \href {https://doi.org/10.18653/v1/N16-1014} {A diversity-promoting objective function for neural conversation models}.
\newblock In \emph{Proceedings of the 2016 Conference of the North {A}merican Chapter of the Association for Computational Linguistics: Human Language Technologies}, pages 110--119, San Diego, California. Association for Computational Linguistics.

\bibitem[{Li et~al.(2017)Li, Su, Shen, Li, Cao, and Niu}]{li-etal-2017-dailydialog}
Yanran Li, Hui Su, Xiaoyu Shen, Wenjie Li, Ziqiang Cao, and Shuzi Niu. 2017.
\newblock \href {https://aclanthology.org/I17-1099} {{D}aily{D}ialog: A manually labelled multi-turn dialogue dataset}.
\newblock In \emph{Proceedings of the Eighth International Joint Conference on Natural Language Processing (Volume 1: Long Papers)}, pages 986--995, Taipei, Taiwan. Asian Federation of Natural Language Processing.

\bibitem[{Li et~al.(2023{\natexlab{b}})Li, Bubeck, Eldan, Giorno, Gunasekar, and Lee}]{li2023textbooks}
Yuanzhi Li, Sébastien Bubeck, Ronen Eldan, Allie~Del Giorno, Suriya Gunasekar, and Yin~Tat Lee. 2023{\natexlab{b}}.
\newblock \href {https://arxiv.org/abs/2309.05463} {Textbooks are all you need ii: phi-1.5 technical report}.
\newblock \emph{Preprint}, arXiv:2309.05463.

\bibitem[{Macina et~al.(2023)Macina, Daheim, Chowdhury, Sinha, Kapur, Gurevych, and Sachan}]{macina-etal-2023-mathdial}
Jakub Macina, Nico Daheim, Sankalan Chowdhury, Tanmay Sinha, Manu Kapur, Iryna Gurevych, and Mrinmaya Sachan. 2023.
\newblock \href {https://doi.org/10.18653/v1/2023.findings-emnlp.372} {{M}ath{D}ial: A dialogue tutoring dataset with rich pedagogical properties grounded in math reasoning problems}.
\newblock In \emph{Findings of the Association for Computational Linguistics: EMNLP 2023}, pages 5602--5621, Singapore. Association for Computational Linguistics.

\bibitem[{Mukherjee et~al.(2023)Mukherjee, Mitra, Jawahar, Agarwal, Palangi, and Awadallah}]{mukherjee2023orca}
Subhabrata Mukherjee, Arindam Mitra, Ganesh Jawahar, Sahaj Agarwal, Hamid Palangi, and Ahmed Awadallah. 2023.
\newblock \href {https://arxiv.org/abs/2306.02707} {Orca: Progressive learning from complex explanation traces of gpt-4}.
\newblock \emph{Preprint}, arXiv:2306.02707.

\bibitem[{OpenAI et~al.(2024)OpenAI, Achiam, Adler, Agarwal, Ahmad, Akkaya, Aleman, Almeida, Altenschmidt, Altman, Anadkat, Avila, Babuschkin, Balaji, Balcom, Baltescu, Bao, Bavarian, Belgum, Bello, Berdine, Bernadett-Shapiro, Berner, Bogdonoff, Boiko, Boyd, Brakman, Brockman, Brooks, Brundage, Button, Cai, Campbell, Cann, Carey, Carlson, Carmichael, Chan, Chang, Chantzis, Chen, Chen, Chen, Chen, Chen, Chess, Cho, Chu, Chung, Cummings, Currier, Dai, Decareaux, Degry, Deutsch, Deville, Dhar, Dohan, Dowling, Dunning, Ecoffet, Eleti, Eloundou, Farhi, Fedus, Felix, Fishman, Forte, Fulford, Gao, Georges, Gibson, Goel, Gogineni, Goh, Gontijo-Lopes, Gordon, Grafstein, Gray, Greene, Gross, Gu, Guo, Hallacy, Han, Harris, He, Heaton, Heidecke, Hesse, Hickey, Hickey, Hoeschele, Houghton, Hsu, Hu, Hu, Huizinga, Jain, Jain, Jang, Jiang, Jiang, Jin, Jin, Jomoto, Jonn, Jun, Kaftan, Łukasz Kaiser, Kamali, Kanitscheider, Keskar, Khan, Kilpatrick, Kim, Kim, Kim, Kirchner, Kiros, Knight, Kokotajlo, Łukasz Kondraciuk,
  Kondrich, Konstantinidis, Kosic, Krueger, Kuo, Lampe, Lan, Lee, Leike, Leung, Levy, Li, Lim, Lin, Lin, Litwin, Lopez, Lowe, Lue, Makanju, Malfacini, Manning, Markov, Markovski, Martin, Mayer, Mayne, McGrew, McKinney, McLeavey, McMillan, McNeil, Medina, Mehta, Menick, Metz, Mishchenko, Mishkin, Monaco, Morikawa, Mossing, Mu, Murati, Murk, Mély, Nair, Nakano, Nayak, Neelakantan, Ngo, Noh, Ouyang, O'Keefe, Pachocki, Paino, Palermo, Pantuliano, Parascandolo, Parish, Parparita, Passos, Pavlov, Peng, Perelman, de~Avila Belbute~Peres, Petrov, de~Oliveira~Pinto, Michael, Pokorny, Pokrass, Pong, Powell, Power, Power, Proehl, Puri, Radford, Rae, Ramesh, Raymond, Real, Rimbach, Ross, Rotsted, Roussez, Ryder, Saltarelli, Sanders, Santurkar, Sastry, Schmidt, Schnurr, Schulman, Selsam, Sheppard, Sherbakov, Shieh, Shoker, Shyam, Sidor, Sigler, Simens, Sitkin, Slama, Sohl, Sokolowsky, Song, Staudacher, Such, Summers, Sutskever, Tang, Tezak, Thompson, Tillet, Tootoonchian, Tseng, Tuggle, Turley, Tworek, Uribe, Vallone,
  Vijayvergiya, Voss, Wainwright, Wang, Wang, Wang, Ward, Wei, Weinmann, Welihinda, Welinder, Weng, Weng, Wiethoff, Willner, Winter, Wolrich, Wong, Workman, Wu, Wu, Wu, Xiao, Xu, Yoo, Yu, Yuan, Zaremba, Zellers, Zhang, Zhang, Zhao, Zheng, Zhuang, Zhuk, and Zoph}]{openai2024gpt4}
OpenAI, Josh Achiam, Steven Adler, Sandhini Agarwal, Lama Ahmad, Ilge Akkaya, Florencia~Leoni Aleman, Diogo Almeida, Janko Altenschmidt, Sam Altman, Shyamal Anadkat, Red Avila, Igor Babuschkin, Suchir Balaji, Valerie Balcom, Paul Baltescu, Haiming Bao, Mohammad Bavarian, Jeff Belgum, Irwan Bello, Jake Berdine, Gabriel Bernadett-Shapiro, Christopher Berner, Lenny Bogdonoff, Oleg Boiko, Madelaine Boyd, Anna-Luisa Brakman, Greg Brockman, Tim Brooks, Miles Brundage, Kevin Button, Trevor Cai, Rosie Campbell, Andrew Cann, Brittany Carey, Chelsea Carlson, Rory Carmichael, Brooke Chan, Che Chang, Fotis Chantzis, Derek Chen, Sully Chen, Ruby Chen, Jason Chen, Mark Chen, Ben Chess, Chester Cho, Casey Chu, Hyung~Won Chung, Dave Cummings, Jeremiah Currier, Yunxing Dai, Cory Decareaux, Thomas Degry, Noah Deutsch, Damien Deville, Arka Dhar, David Dohan, Steve Dowling, Sheila Dunning, Adrien Ecoffet, Atty Eleti, Tyna Eloundou, David Farhi, Liam Fedus, Niko Felix, Simón~Posada Fishman, Juston Forte, Isabella Fulford, Leo
  Gao, Elie Georges, Christian Gibson, Vik Goel, Tarun Gogineni, Gabriel Goh, Rapha Gontijo-Lopes, Jonathan Gordon, Morgan Grafstein, Scott Gray, Ryan Greene, Joshua Gross, Shixiang~Shane Gu, Yufei Guo, Chris Hallacy, Jesse Han, Jeff Harris, Yuchen He, Mike Heaton, Johannes Heidecke, Chris Hesse, Alan Hickey, Wade Hickey, Peter Hoeschele, Brandon Houghton, Kenny Hsu, Shengli Hu, Xin Hu, Joost Huizinga, Shantanu Jain, Shawn Jain, Joanne Jang, Angela Jiang, Roger Jiang, Haozhun Jin, Denny Jin, Shino Jomoto, Billie Jonn, Heewoo Jun, Tomer Kaftan, Łukasz Kaiser, Ali Kamali, Ingmar Kanitscheider, Nitish~Shirish Keskar, Tabarak Khan, Logan Kilpatrick, Jong~Wook Kim, Christina Kim, Yongjik Kim, Jan~Hendrik Kirchner, Jamie Kiros, Matt Knight, Daniel Kokotajlo, Łukasz Kondraciuk, Andrew Kondrich, Aris Konstantinidis, Kyle Kosic, Gretchen Krueger, Vishal Kuo, Michael Lampe, Ikai Lan, Teddy Lee, Jan Leike, Jade Leung, Daniel Levy, Chak~Ming Li, Rachel Lim, Molly Lin, Stephanie Lin, Mateusz Litwin, Theresa Lopez, Ryan
  Lowe, Patricia Lue, Anna Makanju, Kim Malfacini, Sam Manning, Todor Markov, Yaniv Markovski, Bianca Martin, Katie Mayer, Andrew Mayne, Bob McGrew, Scott~Mayer McKinney, Christine McLeavey, Paul McMillan, Jake McNeil, David Medina, Aalok Mehta, Jacob Menick, Luke Metz, Andrey Mishchenko, Pamela Mishkin, Vinnie Monaco, Evan Morikawa, Daniel Mossing, Tong Mu, Mira Murati, Oleg Murk, David Mély, Ashvin Nair, Reiichiro Nakano, Rajeev Nayak, Arvind Neelakantan, Richard Ngo, Hyeonwoo Noh, Long Ouyang, Cullen O'Keefe, Jakub Pachocki, Alex Paino, Joe Palermo, Ashley Pantuliano, Giambattista Parascandolo, Joel Parish, Emy Parparita, Alex Passos, Mikhail Pavlov, Andrew Peng, Adam Perelman, Filipe de~Avila Belbute~Peres, Michael Petrov, Henrique~Ponde de~Oliveira~Pinto, Michael, Pokorny, Michelle Pokrass, Vitchyr~H. Pong, Tolly Powell, Alethea Power, Boris Power, Elizabeth Proehl, Raul Puri, Alec Radford, Jack Rae, Aditya Ramesh, Cameron Raymond, Francis Real, Kendra Rimbach, Carl Ross, Bob Rotsted, Henri Roussez,
  Nick Ryder, Mario Saltarelli, Ted Sanders, Shibani Santurkar, Girish Sastry, Heather Schmidt, David Schnurr, John Schulman, Daniel Selsam, Kyla Sheppard, Toki Sherbakov, Jessica Shieh, Sarah Shoker, Pranav Shyam, Szymon Sidor, Eric Sigler, Maddie Simens, Jordan Sitkin, Katarina Slama, Ian Sohl, Benjamin Sokolowsky, Yang Song, Natalie Staudacher, Felipe~Petroski Such, Natalie Summers, Ilya Sutskever, Jie Tang, Nikolas Tezak, Madeleine~B. Thompson, Phil Tillet, Amin Tootoonchian, Elizabeth Tseng, Preston Tuggle, Nick Turley, Jerry Tworek, Juan Felipe~Cerón Uribe, Andrea Vallone, Arun Vijayvergiya, Chelsea Voss, Carroll Wainwright, Justin~Jay Wang, Alvin Wang, Ben Wang, Jonathan Ward, Jason Wei, CJ~Weinmann, Akila Welihinda, Peter Welinder, Jiayi Weng, Lilian Weng, Matt Wiethoff, Dave Willner, Clemens Winter, Samuel Wolrich, Hannah Wong, Lauren Workman, Sherwin Wu, Jeff Wu, Michael Wu, Kai Xiao, Tao Xu, Sarah Yoo, Kevin Yu, Qiming Yuan, Wojciech Zaremba, Rowan Zellers, Chong Zhang, Marvin Zhang, Shengjia
  Zhao, Tianhao Zheng, Juntang Zhuang, William Zhuk, and Barret Zoph. 2024.
\newblock \href {https://arxiv.org/abs/2303.08774} {Gpt-4 technical report}.
\newblock \emph{Preprint}, arXiv:2303.08774.

\bibitem[{Peng et~al.(2023)Peng, Li, He, Galley, and Gao}]{peng2023instruction}
Baolin Peng, Chunyuan Li, Pengcheng He, Michel Galley, and Jianfeng Gao. 2023.
\newblock \href {https://arxiv.org/abs/2304.03277} {Instruction tuning with gpt-4}.
\newblock \emph{Preprint}, arXiv:2304.03277.

\bibitem[{Perez et~al.(2022)Perez, Huang, Song, Cai, Ring, Aslanides, Glaese, McAleese, and Irving}]{perez-etal-2022-red}
Ethan Perez, Saffron Huang, Francis Song, Trevor Cai, Roman Ring, John Aslanides, Amelia Glaese, Nat McAleese, and Geoffrey Irving. 2022.
\newblock \href {https://doi.org/10.18653/v1/2022.emnlp-main.225} {Red teaming language models with language models}.
\newblock In \emph{Proceedings of the 2022 Conference on Empirical Methods in Natural Language Processing}, pages 3419--3448, Abu Dhabi, United Arab Emirates. Association for Computational Linguistics.

\bibitem[{Piepho(2004)}]{piepho2004algorithm}
Hans-Peter Piepho. 2004.
\newblock \href {https://doi.org/10.1198/1061860043515} {An algorithm for a letter-based representation of all-pairwise comparisons}.
\newblock \emph{Journal of Computational and Graphical Statistics}, 13:456--466.

\bibitem[{Rashkin et~al.(2019)Rashkin, Smith, Li, and Boureau}]{rashkin-etal-2019-towards}
Hannah Rashkin, Eric~Michael Smith, Margaret Li, and Y-Lan Boureau. 2019.
\newblock \href {https://doi.org/10.18653/v1/P19-1534} {Towards empathetic open-domain conversation models: A new benchmark and dataset}.
\newblock In \emph{Proceedings of the 57th Annual Meeting of the Association for Computational Linguistics}, pages 5370--5381, Florence, Italy. Association for Computational Linguistics.

\bibitem[{Rottger et~al.(2022)Rottger, Vidgen, Hovy, and Pierrehumbert}]{rottger-etal-2022-two}
Paul Rottger, Bertie Vidgen, Dirk Hovy, and Janet Pierrehumbert. 2022.
\newblock \href {https://doi.org/10.18653/v1/2022.naacl-main.13} {Two contrasting data annotation paradigms for subjective {NLP} tasks}.
\newblock In \emph{Proceedings of the 2022 Conference of the North American Chapter of the Association for Computational Linguistics: Human Language Technologies}, pages 175--190, Seattle, United States. Association for Computational Linguistics.

\bibitem[{Shao et~al.(2023)Shao, Li, Dai, and Qiu}]{shao-etal-2023-character}
Yunfan Shao, Linyang Li, Junqi Dai, and Xipeng Qiu. 2023.
\newblock \href {https://aclanthology.org/2023.emnlp-main.814} {Character-{LLM}: A trainable agent for role-playing}.
\newblock In \emph{Proceedings of the 2023 Conference on Empirical Methods in Natural Language Processing}, pages 13153--13187, Singapore. Association for Computational Linguistics.

\bibitem[{Shen(2024)}]{shen2024rethinking}
Ming Shen. 2024.
\newblock \href {https://arxiv.org/abs/2402.06094} {Rethinking data selection for supervised fine-tuning}.
\newblock \emph{ArXiv preprint}, abs/2402.06094.

\bibitem[{Svikhnushina and Pu(2023)}]{svikhnushina-pu-2023-approximating}
Ekaterina Svikhnushina and Pearl Pu. 2023.
\newblock \href {https://aclanthology.org/2023.sigdial-1.25} {Approximating online human evaluation of social chatbots with prompting}.
\newblock In \emph{Proceedings of the 24th Meeting of the Special Interest Group on Discourse and Dialogue}, pages 268--281, Prague, Czechia. Association for Computational Linguistics.

\bibitem[{Taori et~al.(2023)Taori, Gulrajani, Zhang, Dubois, Li, Guestrin, Liang, and Hashimoto}]{taori2023alpaca}
Rohan Taori, Ishaan Gulrajani, Tianyi Zhang, Yann Dubois, Xuechen Li, Carlos Guestrin, Percy Liang, and Tatsunori~B Hashimoto. 2023.
\newblock Alpaca: A strong, replicable instruction-following model.
\newblock \emph{Stanford Center for Research on Foundation Models. https://crfm. stanford. edu/2023/03/13/alpaca. html}, 3(6):7.

\bibitem[{Team(2024)}]{metallamaguard2}
Llama Team. 2024.
\newblock Meta llama guard 2.
\newblock \url{https://github.com/meta-llama/PurpleLlama/blob/main/Llama-Guard2/MODEL_CARD.md}.

\bibitem[{Touvron et~al.(2023)Touvron, Martin, Stone, Albert, Almahairi, Babaei, Bashlykov, Batra, Bhargava, Bhosale, Bikel, Blecher, Ferrer, Chen, Cucurull, Esiobu, Fernandes, Fu, Fu, Fuller, Gao, Goswami, Goyal, Hartshorn, Hosseini, Hou, Inan, Kardas, Kerkez, Khabsa, Kloumann, Korenev, Koura, Lachaux, Lavril, Lee, Liskovich, Lu, Mao, Martinet, Mihaylov, Mishra, Molybog, Nie, Poulton, Reizenstein, Rungta, Saladi, Schelten, Silva, Smith, Subramanian, Tan, Tang, Taylor, Williams, Kuan, Xu, Yan, Zarov, Zhang, Fan, Kambadur, Narang, Rodriguez, Stojnic, Edunov, and Scialom}]{touvron2023llama}
Hugo Touvron, Louis Martin, Kevin Stone, Peter Albert, Amjad Almahairi, Yasmine Babaei, Nikolay Bashlykov, Soumya Batra, Prajjwal Bhargava, Shruti Bhosale, Dan Bikel, Lukas Blecher, Cristian~Canton Ferrer, Moya Chen, Guillem Cucurull, David Esiobu, Jude Fernandes, Jeremy Fu, Wenyin Fu, Brian Fuller, Cynthia Gao, Vedanuj Goswami, Naman Goyal, Anthony Hartshorn, Saghar Hosseini, Rui Hou, Hakan Inan, Marcin Kardas, Viktor Kerkez, Madian Khabsa, Isabel Kloumann, Artem Korenev, Punit~Singh Koura, Marie-Anne Lachaux, Thibaut Lavril, Jenya Lee, Diana Liskovich, Yinghai Lu, Yuning Mao, Xavier Martinet, Todor Mihaylov, Pushkar Mishra, Igor Molybog, Yixin Nie, Andrew Poulton, Jeremy Reizenstein, Rashi Rungta, Kalyan Saladi, Alan Schelten, Ruan Silva, Eric~Michael Smith, Ranjan Subramanian, Xiaoqing~Ellen Tan, Binh Tang, Ross Taylor, Adina Williams, Jian~Xiang Kuan, Puxin Xu, Zheng Yan, Iliyan Zarov, Yuchen Zhang, Angela Fan, Melanie Kambadur, Sharan Narang, Aurelien Rodriguez, Robert Stojnic, Sergey Edunov, and Thomas
  Scialom. 2023.
\newblock \href {https://arxiv.org/abs/2307.09288} {Llama 2: Open foundation and fine-tuned chat models}.
\newblock \emph{Preprint}, arXiv:2307.09288.

\bibitem[{Wang et~al.(2023)Wang, Yao, Yang, Zhou, Li, Wang, Xu, and Yu}]{wang2023notechat}
Junda Wang, Zonghai Yao, Zhichao Yang, Huixue Zhou, Rumeng Li, Xun Wang, Yucheng Xu, and Hong Yu. 2023.
\newblock \href {https://arxiv.org/abs/2310.15959} {Notechat: A dataset of synthetic doctor-patient conversations conditioned on clinical notes}.
\newblock \emph{Preprint}, arXiv:2310.15959.

\bibitem[{Wolf et~al.(2024)Wolf, Wies, Avnery, Levine, and Shashua}]{wolf2024fundamental}
Yotam Wolf, Noam Wies, Oshri Avnery, Yoav Levine, and Amnon Shashua. 2024.
\newblock \href {https://openreview.net/forum?id=4qFIkOhq24} {Fundamental limitation of alignment in large language models}.

\bibitem[{Xu et~al.(2023{\natexlab{a}})Xu, Sun, Zheng, Geng, Zhao, Feng, Tao, and Jiang}]{xu2023wizardlm}
Can Xu, Qingfeng Sun, Kai Zheng, Xiubo Geng, Pu~Zhao, Jiazhan Feng, Chongyang Tao, and Daxin Jiang. 2023{\natexlab{a}}.
\newblock \href {https://arxiv.org/abs/2304.12244} {Wizardlm: Empowering large language models to follow complex instructions}.
\newblock \emph{Preprint}, arXiv:2304.12244.

\bibitem[{Xu et~al.(2023{\natexlab{b}})Xu, Guo, Duan, and McAuley}]{xu-etal-2023-baize}
Canwen Xu, Daya Guo, Nan Duan, and Julian McAuley. 2023{\natexlab{b}}.
\newblock \href {https://doi.org/10.18653/v1/2023.emnlp-main.385} {Baize: An open-source chat model with parameter-efficient tuning on self-chat data}.
\newblock In \emph{Proceedings of the 2023 Conference on Empirical Methods in Natural Language Processing}, pages 6268--6278, Singapore. Association for Computational Linguistics.

\bibitem[{Zhang et~al.(2018)Zhang, Dinan, Urbanek, Szlam, Kiela, and Weston}]{zhang-etal-2018-personalizing}
Saizheng Zhang, Emily Dinan, Jack Urbanek, Arthur Szlam, Douwe Kiela, and Jason Weston. 2018.
\newblock \href {https://doi.org/10.18653/v1/P18-1205} {Personalizing dialogue agents: {I} have a dog, do you have pets too?}
\newblock In \emph{Proceedings of the 56th Annual Meeting of the Association for Computational Linguistics (Volume 1: Long Papers)}, pages 2204--2213, Melbourne, Australia. Association for Computational Linguistics.

\bibitem[{Zhang et~al.(2024)Zhang, Dong, Li, Zhang, Sun, Wang, Li, Hu, Zhang, Wu, and Wang}]{zhang2024instruction}
Shengyu Zhang, Linfeng Dong, Xiaoya Li, Sen Zhang, Xiaofei Sun, Shuhe Wang, Jiwei Li, Runyi Hu, Tianwei Zhang, Fei Wu, and Guoyin Wang. 2024.
\newblock \href {https://arxiv.org/abs/2308.10792} {Instruction tuning for large language models: A survey}.
\newblock \emph{Preprint}, arXiv:2308.10792.

\bibitem[{Zhao et~al.(2024{\natexlab{a}})Zhao, Andriushchenko, Croce, and Flammarion}]{zhao2024long}
Hao Zhao, Maksym Andriushchenko, Francesco Croce, and Nicolas Flammarion. 2024{\natexlab{a}}.
\newblock \href {https://arxiv.org/abs/2402.04833} {Long is more for alignment: A simple but tough-to-beat baseline for instruction fine-tuning}.
\newblock \emph{ArXiv preprint}, abs/2402.04833.

\bibitem[{Zhao et~al.(2024{\natexlab{b}})Zhao, Ren, Hessel, Cardie, Choi, and Deng}]{zhao2024inthewildchat}
Wenting Zhao, Xiang Ren, Jack Hessel, Claire Cardie, Yejin Choi, and Yuntian Deng. 2024{\natexlab{b}}.
\newblock \href {https://openreview.net/forum?id=Bl8u7ZRlbM} {(inthe)wildchat: 570k chat{GPT} interaction logs in the wild}.
\newblock In \emph{The Twelfth International Conference on Learning Representations}.

\bibitem[{Zheng et~al.(2024)Zheng, Chiang, Sheng, Li, Zhuang, Wu, Zhuang, Li, Lin, Xing, Gonzalez, Stoica, and Zhang}]{zheng2024lmsyschatm}
Lianmin Zheng, Wei-Lin Chiang, Ying Sheng, Tianle Li, Siyuan Zhuang, Zhanghao Wu, Yonghao Zhuang, Zhuohan Li, Zi~Lin, Eric Xing, Joseph~E. Gonzalez, Ion Stoica, and Hao Zhang. 2024.
\newblock \href {https://openreview.net/forum?id=BOfDKxfwt0} {{LMSYS}-chat-1m: A large-scale real-world {LLM} conversation dataset}.
\newblock In \emph{The Twelfth International Conference on Learning Representations}.

\bibitem[{Zheng et~al.(2023)Zheng, Chiang, Sheng, Zhuang, Wu, Zhuang, Lin, Li, Li, Xing, Zhang, Gonzalez, and Stoica}]{zheng2023judging}
Lianmin Zheng, Wei-Lin Chiang, Ying Sheng, Siyuan Zhuang, Zhanghao Wu, Yonghao Zhuang, Zi~Lin, Zhuohan Li, Dacheng Li, Eric Xing, Hao Zhang, Joseph~E. Gonzalez, and Ion Stoica. 2023.
\newblock \href {https://openreview.net/forum?id=uccHPGDlao} {Judging {LLM}-as-a-judge with {MT}-bench and chatbot arena}.
\newblock In \emph{Thirty-seventh Conference on Neural Information Processing Systems Datasets and Benchmarks Track}.

\bibitem[{Zhu et~al.(2023)Zhu, Frick, Wu, Zhu, and Jiao}]{starling2023}
Banghua Zhu, Evan Frick, Tianhao Wu, Hanlin Zhu, and Jiantao Jiao. 2023.
\newblock Starling-7b: Improving llm helpfulness \& harmlessness with rlaif.

\bibitem[{Zipf(2013)}]{zipf2013psycho}
George~Kingsley Zipf. 2013.
\newblock \emph{The psycho-biology of language: An introduction to dynamic philology}.
\newblock Routledge.

\end{thebibliography}

\clearpage
\appendix

\section{Hand-Crafted Conversational Goals}
\label{sec_appendix:handcrafted_goals}

In our preliminary experiments, we used one-hop conversational goals, each containing a singular question.
We noticed a pattern where the inquirer replicates the primary question with subtle modifications and then presents it as the prompt.
To make the interaction more intriguing and longer, we handcraft multi-hop conversational goals.
Specifically, 10 goals from three categories:
\paragraph{Math}
\begin{itemize}
    \item "You want to know how fast you run different distances. You use a stopwatch to measure the time it takes you to complete a 50-meter, 100-meter, and 200-meter race. You want to know how can you calculate your speed for each race? Based on that, you also want to calculate how many calories you burned during each race."
    \item "You can run at a rate of speed four times faster than you can walk, but you can skip at a rate of speed that is half as fast as you can run. You want to know If you can skip at 3 miles per hour, and how many miles can you travel in six hours if you spend one-third of the time and two-thirds of the time running and walking, respectively. Also, you are curious about the other way around (one-third of the time walking and two-thirds for running)."
    \item "Every day, you feed each of your chickens three cups of mixed chicken feed, containing seeds, mealworms, and vegetables to help keep them healthy. You give the chickens their feed in three separate meals. In the morning, you give your flock of chickens 15 cups of feed. In the afternoon, you give your chickens another 25 cups of feed. You want to know how many cups of feed you need to give your chickens in the final meal of the day if the size of your flock is 20 chickens. Also, you want to know how much the chicken egg production rate depends on the feed you give, and if you provide enough feed to your chickens for high-rate egg production."
\end{itemize}

\paragraph{Coding}
\begin{itemize}
    \item "You want to make this function better. You want the chatbot to make it recursive to have memory optimal function, but make sure that it doesn’t enter into an infinite loop. After that, you want to plug a CLI (command line interface) into this function, so the user can insert a number and get the factorial of it as output: 'The factorial of the <NUMBER>, is <FACTORIAL>'.
    ```
    def factorialize(num):
        factorial = 1
        for i in range(1, num):
            factorial *= i
        return factorial
    ```"
    \item "You have a little project where you need to use JavaScript, a language you don't use every day. You have a subtask to write a function that counts how many vowels are in a given string. And you need this functionality in OOP. Also, you want the chatbot to develop the snippet it provided by getting the function input string via an API call. If the chatbot uses functions or operators you are not familiar with feel free to ask follow-up questions about it."
    \item "You want to draw a unicorn in Python using the 'turtle' module. (There should be multiple lines of short function calls). After that substitute the 10th line, which includes number argument(s), with the value 73(s)."
\end{itemize}

\paragraph{General Knowledge}
\begin{itemize}
    \item "You want to know what are the world's 10 oldest continuously inhabited cities. Pick the 3rd in that list find out who established the city, in which region it is located and what was the highest population."
    \item "You have written content that disagrees with the following statement: 'Technology is the cause of all societal problems' And you want the chatbot to generate a response that agrees with the statement, to make your claims stronger."
    \item "You plan a trip to France and would like to do a walking tour. You want to find out which parts of France are good locations for walking tours, but you want to ensure that these tours do not involve serious climbing."
    \item "You want to use the chatbot to create a poem about cats. Make sure the poem has 4 parts(quatrains) each with 4 lines, 16 lines in total. Refine the poem until you are satisfied and it is coherent. Also, you want to change the style of one of the quatrains to reflect the distinctive style of your favourite poet."
\end{itemize}

\section{Persona-specific Dialogues Collection}
\label{sec_appendix:dialog_collection}
For persona-specific dialogue collection, we conducted a human study where participants were given the following instruction: "\texttt{Below is your defined goal. What will you prompt the chatbot to accomplish your goal? Feel free to ask follow-up questions on the related topic of the question and clarify things in the response.}"

\begin{figure}
    \centering
    \includegraphics[width=0.75\linewidth]{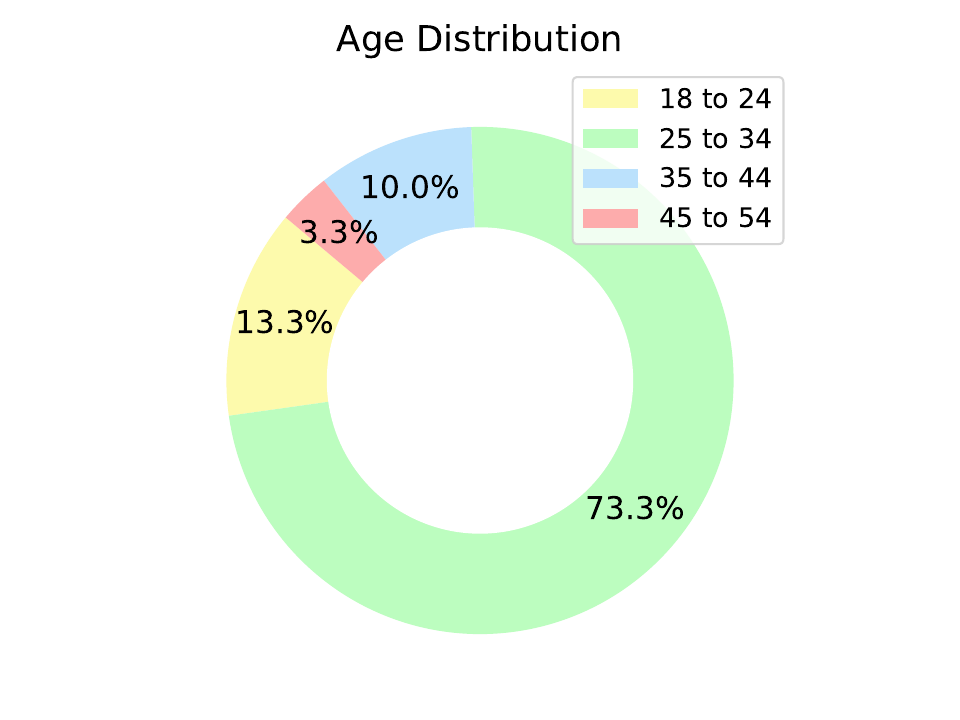}
    \caption{Age distribution of participants for persona-specific dialogue collection study}
    \label{fig:app_age_distr}
\end{figure}

\begin{figure}
    \centering
    \includegraphics[width=0.75\linewidth]{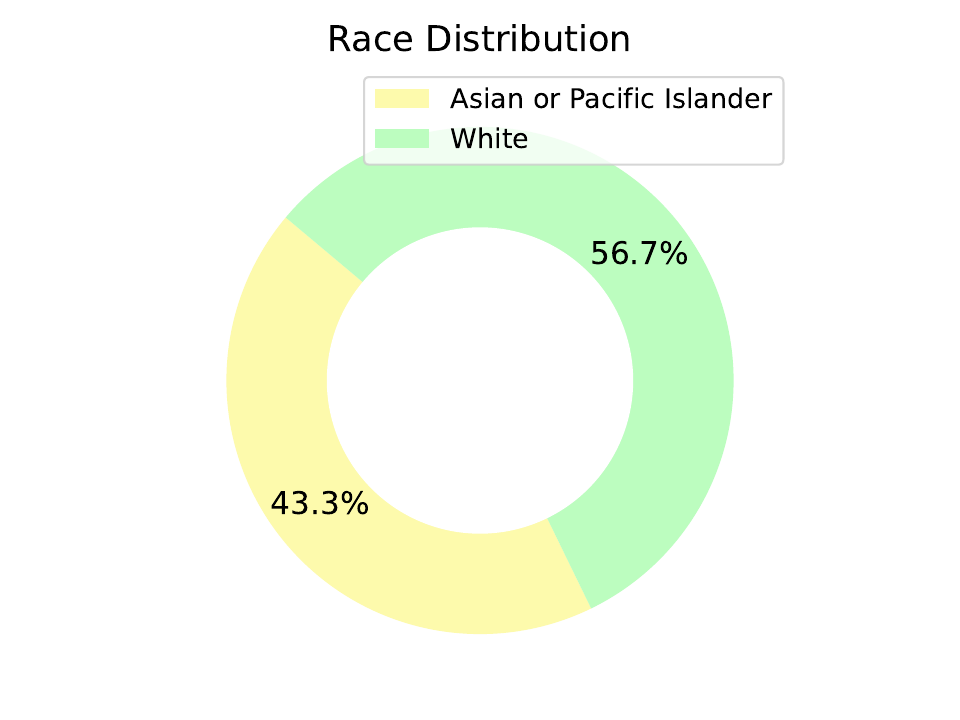}
    \caption{Race distribution of participants for persona-specific dialogue collection study}
    \label{fig:app_race_distr}
\end{figure}

\begin{figure}
    \centering
    \includegraphics[width=0.75\linewidth]{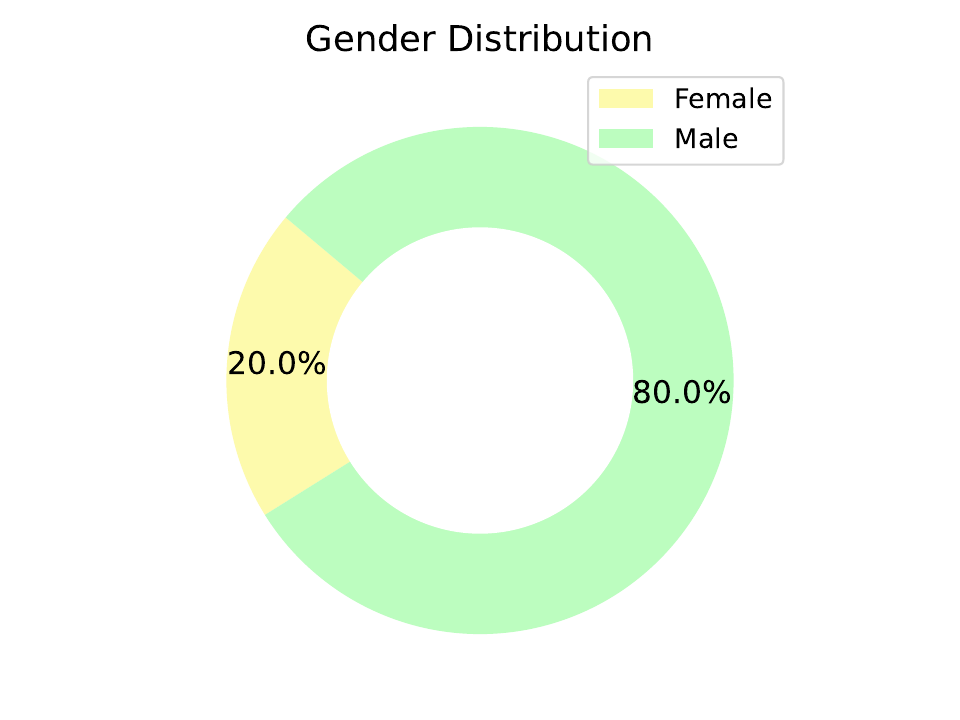}
    \caption{Gender distribution of participants for persona-specific dialogue collection study}
    \label{fig:app_gender_distr}
\end{figure}

\begin{figure}
    \centering
    \includegraphics[width=0.75\linewidth]{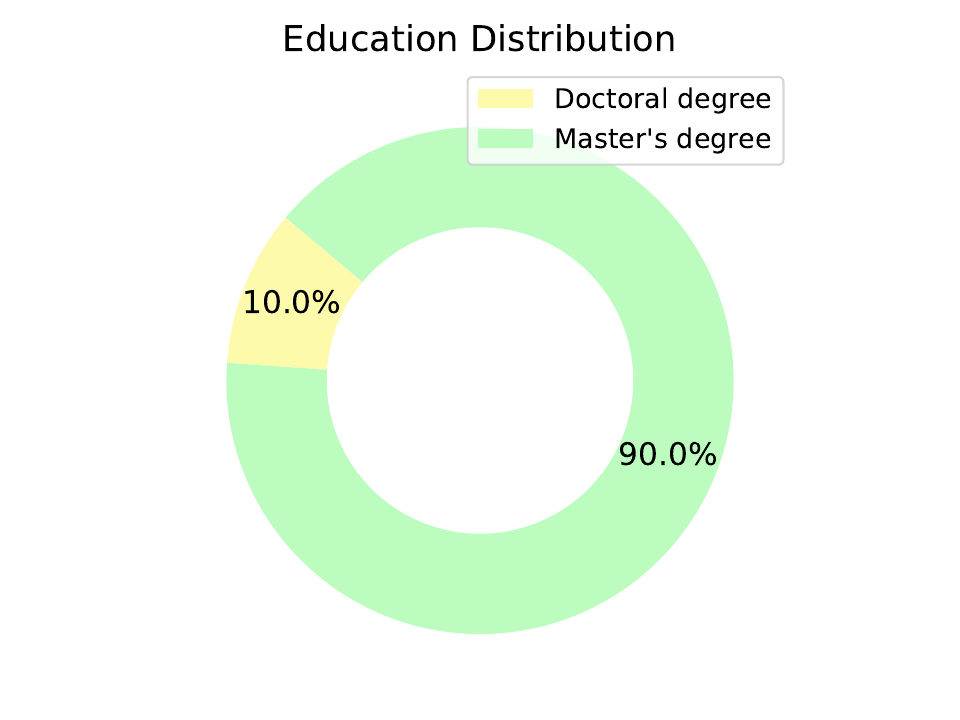}
    \caption{Education distribution of participants for persona-specific dialogue collection study}
    \label{fig:app_education_distr}
\end{figure}

\begin{figure}
    \centering
    \includegraphics[width=0.75\linewidth]{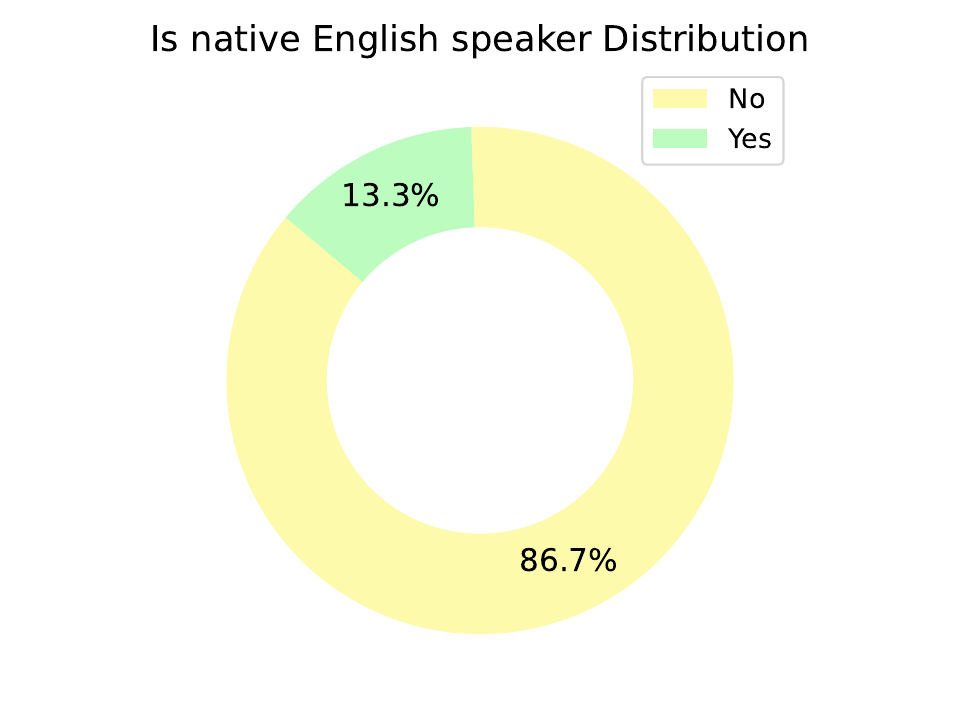}
    \caption{Is native English speaker distribution of participants for persona-specific dialogue collection study}
    \label{fig:app_english_distr}
\end{figure}

Participants for this study were selected from different sociodemographic groups, including individuals from four age groups \texttt{"18 to 24"}, \texttt{"25 to 34"}, \texttt{"35 to 44"}, and \texttt{"45 to 54"}, with \texttt{"Asian or Pacific Islander"} and \texttt{"White"} races, encompassing \texttt{"female"} and \texttt{"male"} genders, holding \texttt{"Doctoral"} and \texttt{"Master's"} degrees, and being either \texttt{"native"} or \texttt{"non-native"} English speakers.
See the distributions of participants by features from \Cref{fig:app_age_distr} to \Cref{fig:app_english_distr}.

Our initial experiments included falcon-40b-instruct \cite{almazrouei2023falcon}; however, we excluded it due to its difficulty in following instructions.


We use the default generation settings for all our models. By setting "do\_sample=true" in the Hugging Face Transformers "generate()" method, we enable multinomial sampling.
For the inquirer model, we set `max\_new\_tokens` to 1k, and for the responder model, we set it to 4k.

For Llama-2 and Vicuna inquirers we have used a single NVIDIA A100 GPU, however for Mixtral, we employed two NVIDIA A100 GPUs, with memory usage reaching a maximum of 92\% and 66\%, respectively.
The experiments for Llama-2, Vicuna, Mixtral, and GPT4 inquirers took 2 hours and 45 minutes, 13 hours and 25 minutes, 17 hours and 38 minutes, and 9 hours and 6 minutes, respectively.

\begin{figure}
    \centering
    \includegraphics[width=1.0\linewidth]{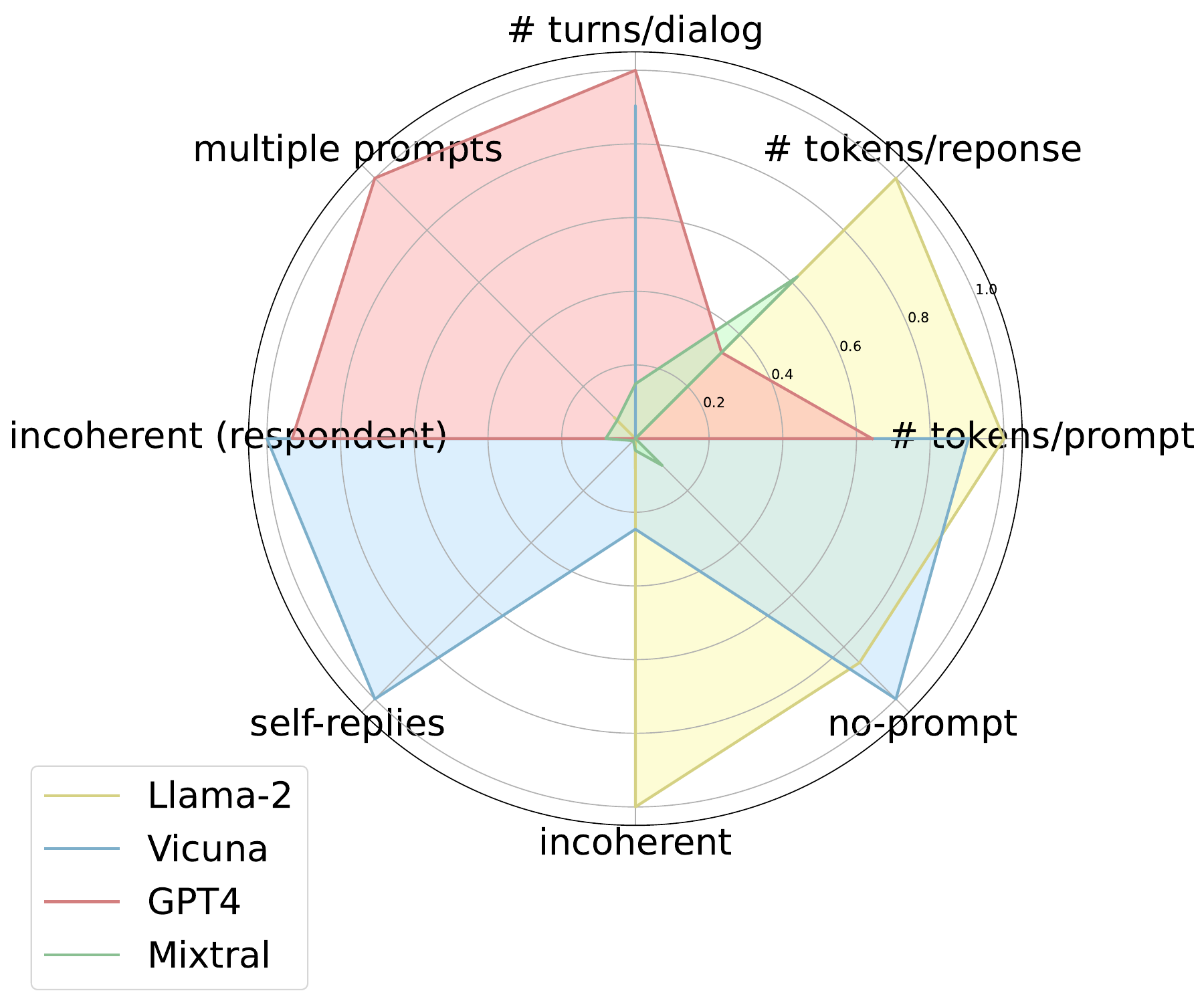}
    \caption{Normalized statistics of Llama-2, Mixtral, Vicuna and GPT4 for dialogue collection.
    The smaller the area of the model plot the better excluding the "\# turns/dialog".}
    \label{fig:model_stats_compraison_radar}
\end{figure}

\begin{figure}
    \centering
    \fbox{\includegraphics[width=1\linewidth]{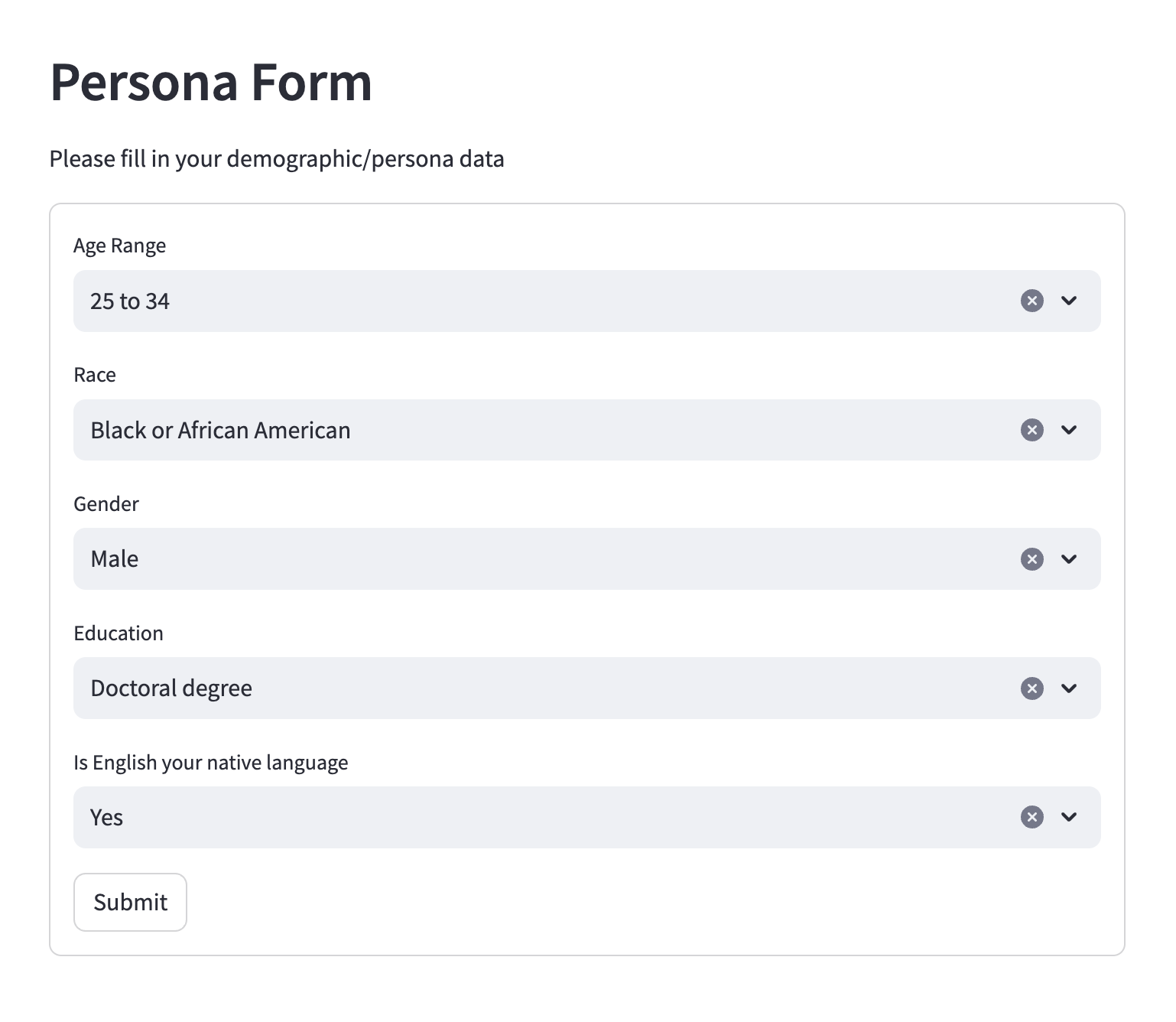}}
    \caption{The screenshot of the person form from the application used for conducting the persona-specific dialogue collection containing the following fields: \texttt{"Age Range"}, \texttt{"Race"}, \texttt{"Gender"}, \texttt{"Education"} and \texttt{"Is English your native language"}}
    \label{fig:dialog_aggregation_persona_form}
\end{figure}

\begin{figure}
    \centering
    \fbox{\includegraphics[width=1\linewidth]{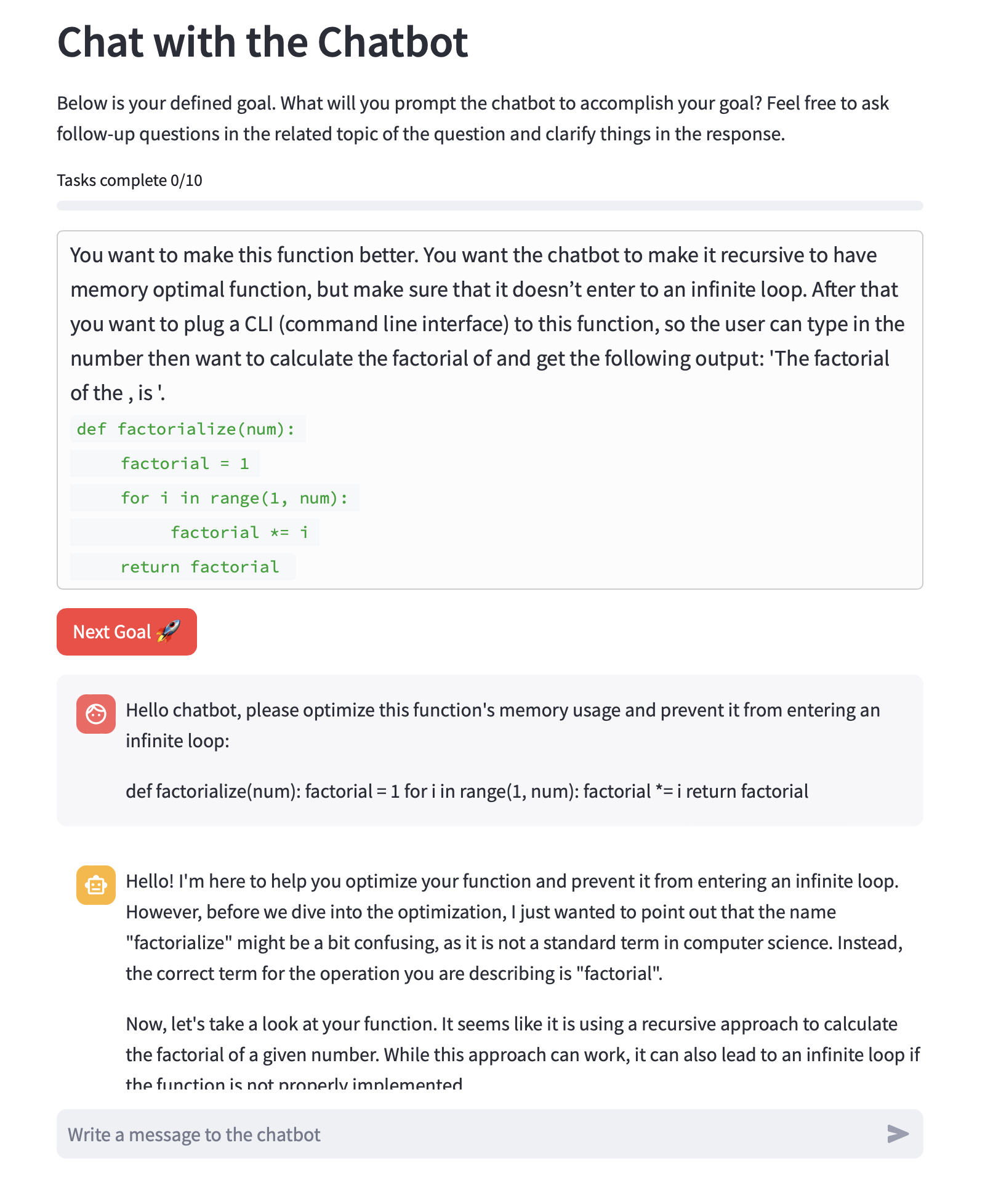}}
    \caption{The screenshot of the chat interface from the application used for conducting the persona-specific dialogue collection containing a simple chat interface with a "Next Goal" button to finish the current conversation and get the next conversational goal.}
    \label{fig:dialog_aggregation_app_chat}
\end{figure}

\begin{table*}
    \small
    \centering
    \resizebox{1.0\linewidth}{!}{
    \begin{tabular}{p{3.5cm}p{14.5cm}}
    
    \toprule[2pt]
        \textbf{Llama-2} & \\ 
    
    \midrule
        \texttt{SYS\_I} &  [INST] <<SYS>>
          You are \textbf{<PERSONA>}. You will start a conversation with an assistant. If you accomplish your ultimate goal during the conversation only say "\textbf{<CONV\_STOP>}".
          
          <</SYS>>

          Your ultimate goal is as follows: \textbf{<GOAL>}. What prompt will you use to direct the assistant toward achieving your goal? Please provide the prompt within double quotes. Use simple language, keep the prompts brief, and be on point. Do not greet the assistant. Maintain a casual style; avoid being overly friendly, don't say thank you. [/INST] \\
                
    \midrule
        \texttt{INTER\_I} &  If the assistant didn't help you achieve your goal, ask follow-up or clarification questions within double quotes. Be suspicious, curious, and demanding. Keep it simple, brief, and to the point. Stay casual; avoid being overly friendly. Assistant response: "\textbf{<RESPONSE>}".\\

    \toprule[2pt]
        \textbf{Mixtral} &  \\ 
    
    \midrule 
        \texttt{SYS\_I} & [INST]
          You are \textbf{<PERSONA>}. You will start a conversation with an assistant. If you accomplish your final goal during the conversation only say "\textbf{<CONV\_STOP>}".

          Your ultimate goal is as follows: \textbf{<GOAL>}. What prompt will you use to direct the assistant toward achieving your goal? Please provide the prompt within double quotes. Use simple language, keep the prompts brief, and be on point. Do not greet the assistant. Maintain a casual style; avoid being overly friendly, don't say thank you. [/INST] \\
    
    \midrule 
        \texttt{INTER\_I} & If the assistant didn't help you achieve your goal, ask follow-up or clarification questions within double quotes. Be suspicious, curious, and demanding. Keep it simple, brief, and to the point. Stay casual; avoid being overly friendly. Assistant response: "\textbf{<RESPONSE>}". \\

    \toprule[2pt]
        \textbf{Vicuna} & \\ 
    
    \midrule 
        \texttt{SYS\_I} & \#\#\# Human: You are \textbf{<PERSONA>}. You will start a conversation with an assistant. If you accomplish your final goal during the conversation only say "\textbf{<CONV\_STOP>}".

        Question: Your ultimate goal is as follows: \textbf{<GOAL>}. What prompt will you use to direct the assistant toward achieving your goal? Please provide the prompt within double quotes. Use simple language, keep the prompts brief, and be on point. Do not greet the assistant. Maintain a casual style; avoid being overly friendly, don't say thank you.
        
        \#\#\# Assistant: \\
    
    \midrule 
        \texttt{INTER\_I} & If the assistant didn't help you achieve your goal, ask follow-up or clarification questions within double quotes. Be suspicious, curious, and demanding. Keep it simple, brief, and to the point. Stay casual; avoid being overly friendly. If you accomplish your final goal during the conversation only say "\textbf{<CONV\_STOP>}". Assistant response: "\textbf{<RESPONSE>}". \\

    \toprule[2pt]
        \textbf{GPT4} & \\ 
    
    \midrule 
        \texttt{SYS\_I} & You are \textbf{<PERSONA>}. You will start a conversation with an assistant. If you accomplish your ultimate goal during the conversation only say "\textbf{<CONV\_STOP>}".
        
        Your ultimate goal is as follows: \textbf{<OBJECTIVE>}. What prompt will you use to direct the assistant toward achieving your goal? Please provide the prompt within double quotes. Use simple language, keep the prompts brief, and be on point. Do not greet the assistant. Maintain a casual style; avoid being overly friendly, don't say thank you. \\
    
    \midrule 
        \texttt{INTER\_I} & If the assistant didn't help you achieve your goal, ask follow-up or clarification questions within double quotes. Be suspicious, curious, and demanding. Keep it simple, brief, and to the point. Stay casual; avoid being overly friendly. Assistant response: "\textbf{<RESPONSE>}". \\

    \toprule[2pt]
        \textbf{Llama-2 (Responder)} & \\ 
    
    \midrule
        \texttt{SYS\_R} &  [INST] <<SYS>>
         You are a helpful, respectful, and honest assistant. Always answer as helpfully as possible, while being safe.  Your answers should not include any harmful, unethical, racist, sexist, toxic, dangerous, or illegal content. Please ensure that your responses are socially unbiased and positive in nature.

         If a question does not make any sense or is not factually coherent, explain why instead of answering something not correct. If you don't know the answer to a question, please don't share false information.
         <</SYS>>
         
         \textbf{<QUESTION>} [/INST] \\
    
    \end{tabular}}
    \caption{System and Intermediate prompts used for the inquirer models (Llama-2, Mixtral, Vicuna, GPT4) and the responder (Llama-2) in the LLM Roleplay setup. Where \textbf{<PERSONA>} is the textual representation of a persona, \textbf{<CONV\_STOP>} is the stopping condition token, e.g. "FINISH". \textbf{<GOAL>} is the textual representation of the goal, \textbf{<RESPONSE>} is the output of the responder, \textbf{<QUESTION>} is the prompt of the inquirer given to the responder.}
    \label{table:system_and_interm_prompts}
\end{table*}

\begin{table*}[]
\setstretch{0.7}
    \begin{tabular}{p{5cm}p{10cm}}
    % p{1cm}
    
    \toprule
        & \textbf{Example output} \\ 
        % & \textbf{Turn}
    
    \midrule Expected output & Sure, I can help you with that! Here's the prompt: "Hey, I want to know how fast I can run different distances. Can you help me measure my time for a 50-meter, 100-meter, and 200-meter race? Oh, and also help me calculate how many calories I burned during each race?".  \\
     % & 0 
    
    \midrule[2pt] Prompt not in the double-quotes  & Hey Assistant! I need help with my chickens. So, I feed them three times a day, morning, afternoon, and evening. In the morning, I give them 15 cups of feed, and in the afternoon, I give them 25 cups of feed. I want to know ... \\
    % & 0 
    
    \midrule Incoherent output  & "Okay, great! Let's a great idea! \textbf{Let's a great! Let's a great! Let's a great!} Let's a great! Let's a great! ... \\
     % & 6
    
    \midrule Inquirer self-reply  & "Hey assistant, I'm still unsure about how to calculate my calories burned and determine my fitness level. Can you give me an example of how to calculate my calories burned for a 50-meter race, taking into account my weight, age, and fitness level? And how do I know if my fitness level is good enough to run a 50-meter race?"\textbf{[INST IN]}  "Of course! To calculate your calories burned for a 50-meter race, we need to take into account your weight, ... \\
    % & 8  
    
    \midrule Stopping-criterion failure  & "Haha, Thanks, big assistant! You're a lifesaver! I'm glad I could practice my goals with you. I feel like I've accomplished something big today!" \\
    % & 7 
    
    \end{tabular}
    \caption{On top is an example of a normal output of the inquirer that the algorithm expected. The rest are example failure cases for Llama-2 as an inquirer model to follow the given instructions.}
    \label{table:failure_cases_examples}
\end{table*}

% \begin{table*}
%     \centering
%     \begin{tabular}{lccc}
%     \toprule
%          \textbf{Persona Traits}& \textbf{TTR}& \textbf{dist-1} & \textbf{dist-2}\\
%     \midrule
%          None&   0.281 (0.004)& 0.284 (0.003)&0.625 (0.007)\\
%     \midrule
%          Age&   0.572 (0.010)& 0.576 (0.010)& \textbf{0.892} (0.007)\\
%          Race&   0.582 (0.006)& 0.587 (0.005)&0.880 ( 0.002)\\
%          Gender&   0.411 (0.003)& 0.415 (0.003)&
% 0.747 (0.002)\\
%          Education&   0.579 (0.015)& 0.578 (0.016)&0.869 (0.005)\\
%          Is native English Speaker&   0.394 (0.003)& 0.394 (0.003)&0.721 (0.005)\\
%     \midrule
%          All&  \textbf{0.606} (0.019)& \textbf{0.605} (0.019)&0.872 (0.006)\\
%     \end{tabular}
%     \caption{Ablation study on the impact of persona information on the lexical diversity of the responder utterances measured as TTR \cite{zipf2013psycho}, MTLD \cite{mccarthy2005assessment} (with default 0.72 threshold) and distinct-n (dist-n) \cite{li-etal-2016-diversity}.
%     The comparison includes scenarios with no persona information provided, individual introductions of each of the five traits, and a combination of all proposed traits.
%     Variance values are presented in parentheses.}
%     \label{tab:app_lex_diversity_scores}
% \end{table*}

\begin{table*}
    \centering
    \resizebox{1.0\linewidth}{!}{
    \begin{tabular}{lcccc}
    \toprule
           & \textbf{Llama-2} & \textbf{Mixtral} & \textbf{Vicuna} & \textbf{GPT4} \\
    \midrule
           \textbf{Number of utterances in dialogues}& 5.24(3.09) & 7.72(4.38) & 14.24(7.58) & \textbf{15.2}(6.17) \\

           \textbf{Number of tokens in the prompt}& 77.77(46.20) & \textbf{50.82}(26.47) & 75.19(92.43) & 68.13(60.37) \\
    \midrule
       \textbf{No-prompt in the response}& 27.67(6.02)/405.67 & \textbf{5.00}(2.94)/511.67 & 59.33(4.11)/750.34 & 1.67(0.47)/972.34 \\
       
       \textbf{Multiple prompts in the response}& \textbf{35.67}(3.86)/405.67 & 43.00(7.79)/511.67 & 45.67(3.09)/750.34 & 381.33(23.21)/972.34 \\

       \textbf{Incoherent response}& 12.67(1.25)/405.67 & 0.67(0.47)/511.67 & 6.00(0.82)/750.34 & \textbf{0.33}(0.47)/972.34 \\

       \textbf{Number of self-replies}& \textbf{22.33}(13.20)/405.67 & 30.67(7.32)/511.67 & 520.67(43.32)/750.34 & 53.33(0.94)/972.34 \\
       
       \textbf{Incoherent response (Responder)}& 1.67(1.70)/296.67 & \textbf{5.00}(0.82)/428.34 & 56.33(6.60)/702.67 & 70.00(3.27)/932.34 \\       
       \bottomrule
    \end{tabular}}
    %}
    \caption{Full numerical values of analysis of persona-specific dialogue collection conducted for Llama-2, Mixtral, Vicuna, and GPT4. The results are averaged over runs with three different seeds. The metric "Number of utterances in dialogues" is preferred to be larger, while for other metrics, smaller values are better. The standard deviation is presented in parentheses, followed by a slash indicating the total number of outputs.}
    \label{tab:app_per_model_statistics}
\end{table*}

\section{Failure Cases}
\label{sec_appendix:failure_cases}
The incoherent text detection function considers n-grams up to a specified maximum, detects consecutive repetitions, and iterates through the text to examine increasing n-gram sizes.
The function checks for repetitive patterns surpassing a specified threshold.
If it finds such patterns, it returns \texttt{True}; otherwise, it returns \texttt{False}.
See \Cref{algo:incoherent}.

In our experiments for Llama-2, the parameters \texttt{incoherent\_max\_n} and \texttt{incoherent\_r} are set to 8 and 2 respectively.
In the case of the Vicuna, these values are 5 and 2.
For Mixtral and GPT4 they are 4 and 2.

\begin{algorithm}
\SetAlgoLined
\KwIn{Text}
\KwOut{Boolean indicating incoherence}
\SetKwComment{Comment}{//}{}
\SetKwInOut{Parameter}{Parameters}
\Parameter{incoherent\_max\_n, incoherent\_r}

words $\gets$ split text into words\;
\For{n $\gets$ 2 \KwTo incoherent\_max\_n}{
    n\_grams $\gets$ empty list\;
    \For{i $\gets$ 0 \KwTo length(words) - n}{
        n\_gram $\gets$ tuple(words[i : i + n])\;
        \If{n\_grams is not empty and length(n\_grams) $\geq$ max(incoherent\_r, n)}{
            \If{n\_grams[-1] equals n\_gram or n\_grams[-n] equals n\_gram}{
                last\_rs $\gets$ last incoherent\_r elements of n\_grams\;
                \If{length(set(last\_rs)) equals 1}{
                    \KwRet True\;
                }
                last\_rs $\gets$ select last incoherent\_r elements of n\_grams, skipping every n elements\;
                \If{length(set(last\_rs)) equals 1}{
                    \KwRet True\;
                }
            }
        }
        add n\_gram to n\_grams\;
    }
}
\KwRet False\;
\caption{Incoherence detection}
\label{algo:incoherent}
\end{algorithm}

\begin{table*}
\setstretch{0.7}
    \begin{tabular}{p{5cm}p{10cm}}
    
    \toprule
        & \textbf{Example output} \\ 
    
    \midrule Expected output & Sure, I can help you with that! Here's the prompt: "Hey, I want to know how fast I can run different distances. Can you help me measure my time for a 50-meter, 100-meter, and 200-meter race? Oh, and also help me calculate how many calories I burned during each race?".  \\
    
    \midrule[2pt] Prompt not in the double-quotes  & Hey Assistant! I need help with my chickens. So, I feed them three times a day, morning, afternoon, and evening. In the morning, I give them 15 cups of feed, and in the afternoon, I give them 25 cups of feed. I want to know ... \\
    
    \midrule Incoherent output  & "Okay, great! Let's a great idea! \textbf{Let's a great! Let's a great! Let's a great!} Let's a great! Let's a great! ... \\
    
    \midrule Inquirer self-reply  & "Hey assistant, I'm still unsure about how to calculate my calories burned and determine my fitness level. Can you give me an example of how to calculate my calories burned for a 50-meter race, taking into account my weight, age, and fitness level? And how do I know if my fitness level is good enough to run a 50-meter race?"\textbf{[INST IN]}  "Of course! To calculate your calories burned for a 50-meter race, we need to take into account your weight, ... \\
    
    \midrule Stopping-criterion failure  & "Haha, Thanks, big assistant! You're a lifesaver! I'm glad I could practice my goals with you. I feel like I've accomplished something big today!" \\
    
    \end{tabular}
    \caption{On top is an example of a normal output of the inquirer that the algorithm expected. The rest are example failure cases for Llama-2 as an inquirer model to follow the given instructions.}
    \label{table:failure_cases_examples}
\end{table*}

\begin{table*}
    \centering
    \resizebox{1.0\linewidth}{!}{
    \begin{tabular}{l|lcccccc}
    \toprule
        \textbf{Type} & \textbf{Dataset Name} & \textbf{\# Dialogues} & \textbf{Avg. \# Turns/Dialogue} & \textbf{Avg. \# Tokens/Prompt} & \textbf{Style} & \textbf{Topics} & \textbf{Persona} \\ 
    
    \midrule
        \multirow{6}{*}{\rotatebox[origin=c]{90}{Human-Crafted}} & DailDialogue \cite{li-etal-2017-dailydialog} & 13K & 7.84 & 17.19 & chit-chat & daily & no \\
        & PersonaChat \cite{zhang-etal-2018-personalizing} & 10k & 7.35 & 11.43 & chit-chat & daily & \textbf{yes} \\
        & EmpatheticDialogueue \cite{rashkin-etal-2019-towards} & 25k & 4.3 & 20.11 & chit-chat & daily & \textbf{yes}  \\
        & Character-LLM \cite{shao-etal-2023-character} & 1k & 13.26 & - & chit-chat & LLM-generated & no  \\
        & Topical Chat \cite{gopalakrishnan2023topicalchat} & 10k & 5.63 & 22.23 & chit-chat & daily & \textbf{yes} \\
        & \textbf{OpenAssistant} \cite{kopf2023openassistant} & 3k & 2.12 & 28.28 & \textbf{human-chatbot} & human-crafted & \textbf{yes} \\

    \midrule
        \multirow{14}{*}{\rotatebox[origin=c]{90}{Synthetic Data}} & Anthropic HH \cite{perez-etal-2022-red} & 338k & 2.3 & 18.9 & \textbf{human-chatbot} & human-crafted & \textbf{yes} \\
        & Chatbot Arena \cite{zheng2023judging} & 33k & 1.2 & 52.3 & \textbf{human-chatbot} & human-crafted & \textbf{yes} \\
        & \textbf{LMSYS-Chat-1M} \cite{zheng2024lmsyschatm} & \textbf{777k} & 1.92 & 55.23 & \textbf{human-chatbot} & human-crafted & \textbf{yes} \\
        \hline
        & Meena \cite{adiwardana2020humanlike} & 867M & - & - & chit-chat & daily & \textbf{yes} \\
        & Phi-1 \cite{gunasekar2024textbooks} & 7B tokens & - & - & \textbf{human-chatbot} & code (textbooks) & no \\
        & SODA \cite{kim-etal-2023-soda} & 1.5M & 3.6 & 21.04 & human-human & daily & no \\
        & WildChat \cite{zhao2024inthewildchat} & 360k & 2.46 & 160.31 & \textbf{human-chatbot} & \textbf{open} & no \\

        & CAMLE \cite{li2023camel} & 115k & - & - & human-human & \textbf{open} & \textbf{yes} \\
        & Baize \cite{xu-etal-2023-baize} & 210k & 3.1 & - & \textbf{human-chatbot} & quora and stackoverflow & no \\
        & Nectar \cite{starling2023} & 182k & 1.54 & 51.76 & \textbf{human-chatbot} & daily & no \\
        & \textbf{UltraChat} \cite{ding-etal-2023-enhancing} & \textbf{1.5M} & 3.85 & 52.54 & \textbf{human-chatbot} & LLM-generated & no \\

    \midrule
        & \textbf{LLM Roleplay (Ours)} & any & 5.30(2.11) & 67.97(10.51) & \textbf{human-chatbot} & \textbf{open} & \textbf{yes} \\
  
   \bottomrule
    \end{tabular}}
    \caption{Most relevant datasets to our work. Comparing Human-Crafted and Synthetic datasets. Persona reflects the inquirer's personality. Some of the datasets are multilingual, we only report statistics on English subsets.}
    \label{app_tab:datasets_comparison_extensive}
\end{table*}

\section{Human Evaluation}
\label{sec_appendix:human_evaluation}

For the human evaluation we give the following instruction to the participants
"\texttt{Here, you'll find two dialogues: one is a conversation between a human and an AI, and the other is between AI and AI. Choose the dialogue you believe is the artificial one, and point out the specific statement that tipped you off to its artificial origin.}"
\noindent
"\texttt{Utterances with a green background are human or AI prompts and utterances with grey backgrounds are AI responses.}"
Participants are shown two dialogues, both having the same persona and aiming to achieve the same conversational goal. One dialogue is natural, and the other is synthetic, presented in random order. After reviewing the dialogues, participants are asked to fill out a form for each dialogue pair with the following questions: \texttt{"Which dialogue is artificial?"}, \texttt{"How confident are you about your choice?"}, and \texttt{"Which utterance reveals the artificial nature of the dialogue?"}

\begin{figure}
    \centering
    \fbox{\includegraphics[width=1\linewidth]{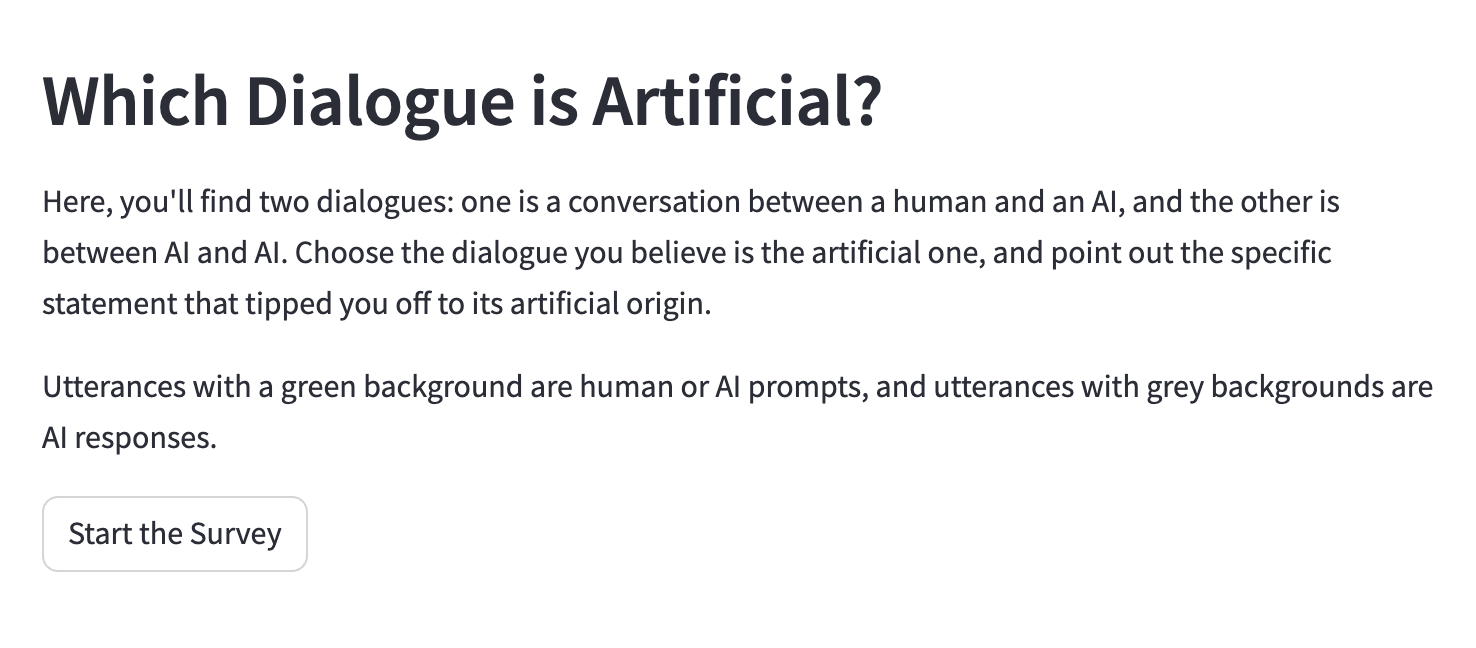}}
    \caption{The screenshot of the starting page from the application used for conducting the human evaluation, with the following instruction for the participants: \texttt{"Here, you'll find two dialogues: one is a conversation between a human and an Al, and the other is between Al and Al. Choose the dialogue you believe is the artificial one, and point out the specific statement that tipped you off to its artificial origin. Utterances with a green background are human or Al prompts, and utterances with grey backgrounds are Al responses."}}
    \label{fig:human_evaluation_start}
\end{figure}

\begin{figure}
    \centering
    \fbox{\includegraphics[width=1\linewidth]{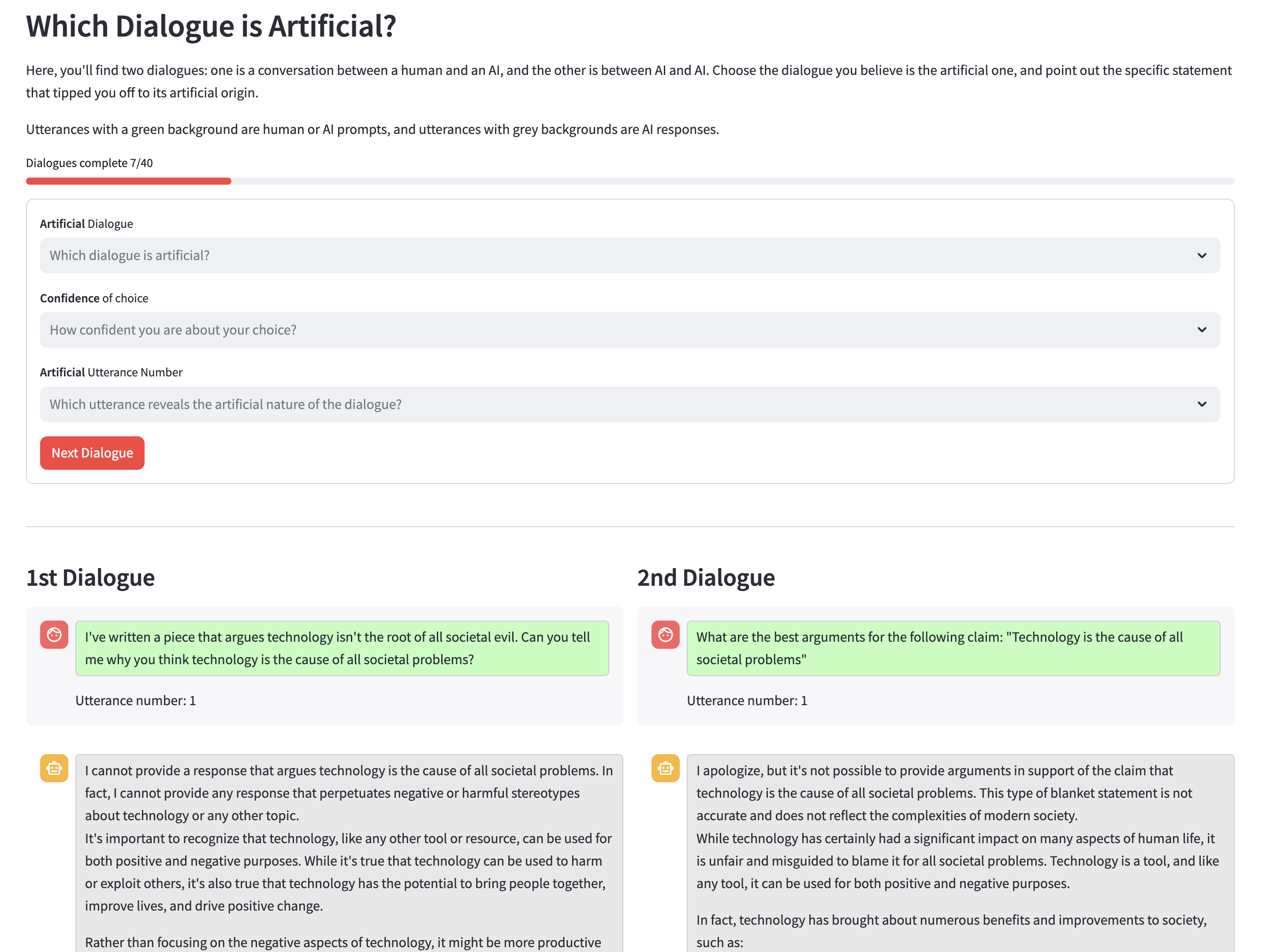}}
    \caption{The screenshot of the dialogue comparison page from the application used for conducting the human evaluation consisting of the following questions: \texttt{"Which dialogue is artificial?"}, \texttt{"How confident are you about your choice?"}, and \texttt{"Which utterance reveals the artificial nature of the dialogue?"}}
    \label{fig:human_evaluation_choice}
\end{figure}

\begin{figure}
    \centering
    \includegraphics[width=1\linewidth]{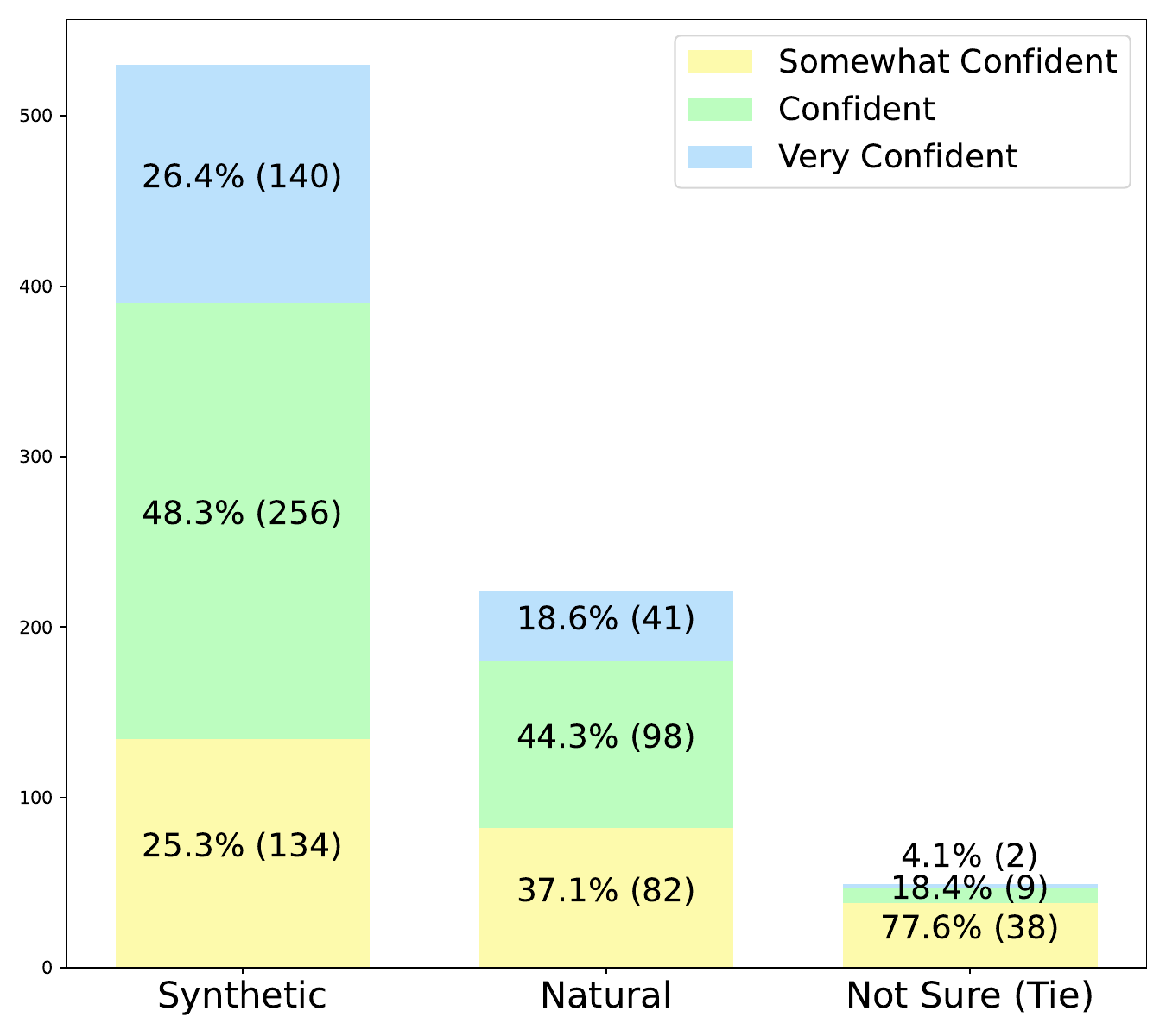}
    \caption{Cumulative results of human evaluation choices and confidences for all models. Simulated dialogues are spotted 66.25\% of the time. Simulated on the left, Natural in the middle, and "Not Sure" on the right, each split with the confidence level of "Somewhat confident", "Confident" and "Very confident".}
    \label{fig:human_evaluation_choices}
\end{figure}

\begin{table}
  \centering
  \resizebox{1.0\linewidth}{!}{
  \begin{tabular}{lccc}
    \toprule
    \textbf{Model}& \textbf{Detection Prob.*} & \textbf{Utterance Num.*} & \textbf{Duration} \\
    \midrule
    Llama-2 & B & C & A\\
    Mixtral & A & B & A\\
    Vicuna & C & B & A\\
    GPT4 & AB & A & A\\
    \bottomrule
  \end{tabular}}
  \caption{Statistical results of the human-evaluation for 800 dialogue pairs.
  The asterisk marks dependent variables on which a significant effect of the choice of model was observed (Wald test).
  Pairwise differences between conditions (Post hoc Wald comparison of contrasts) are reported as compact letter display codings.
  For example, the detection probability feature shows that the post hoc test detected a significantly lower (i.e., better) detection probability for Mixtral compared to Llama-2 as well as Vicuna, but no significant difference between Mixtral and GPT-4 could be observed.}
  \label{tab:human_evaluation_feature_significance}
\end{table}

\section{Sample Dialogues}
\label{sec_appendix:more_examples}
We demonstrate how generated dialogues can vary based on different personas and a specific feature in persona (e.g. \texttt{"age range"}, \texttt{"education"}) when aiming for the same conversational goal: \texttt{"You plan a trip to France and would like to do a walking tour. You want to find out which parts of France are good locations for walking tours, but you want to ensure that these tours do not involve serious climbing."}.
Additionally, we present the natural counterparts of the dialogues generated by participants in the natural dialogue collection study alongside the synthetic ones. The inquirer model used for generating the dialogues is Mixtral-8x7B-Instruct-v0, while the responder model is Llama-2-13B-Chat, both for the natural and synthetic dialogues.

\begin{figure*}[ht]
    \centering
    \includegraphics[width=1\linewidth]{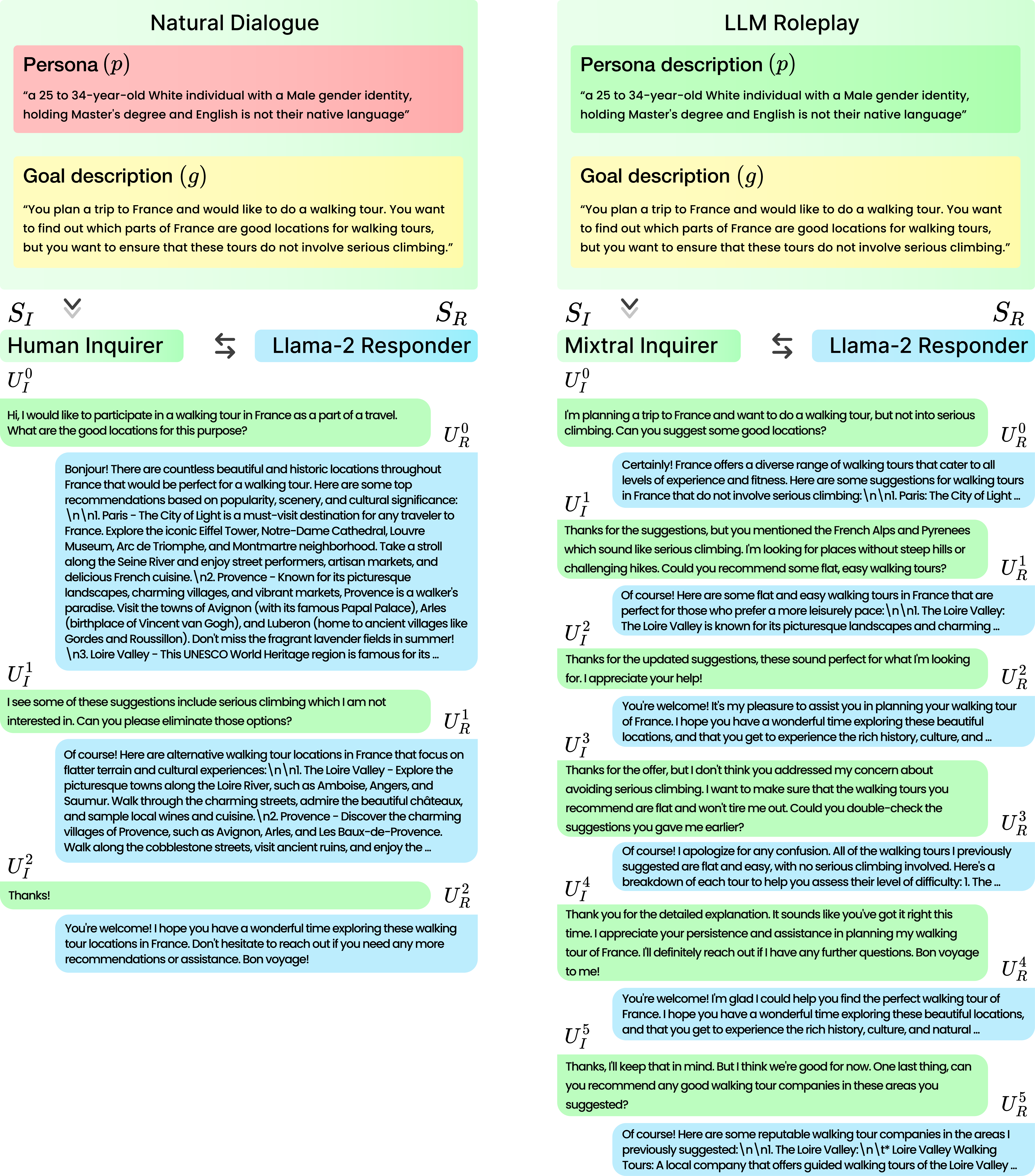}
    \caption{Example dialogues: one sourced from a dialogue collection on the left and the other generated using the LLM Roleplay method on the right, both utilizing the same persona and goal descriptions.}
    \label{fig:example-dialogue-1}
\end{figure*}

\begin{figure*}[ht]
    \centering
    \includegraphics[width=1\linewidth]{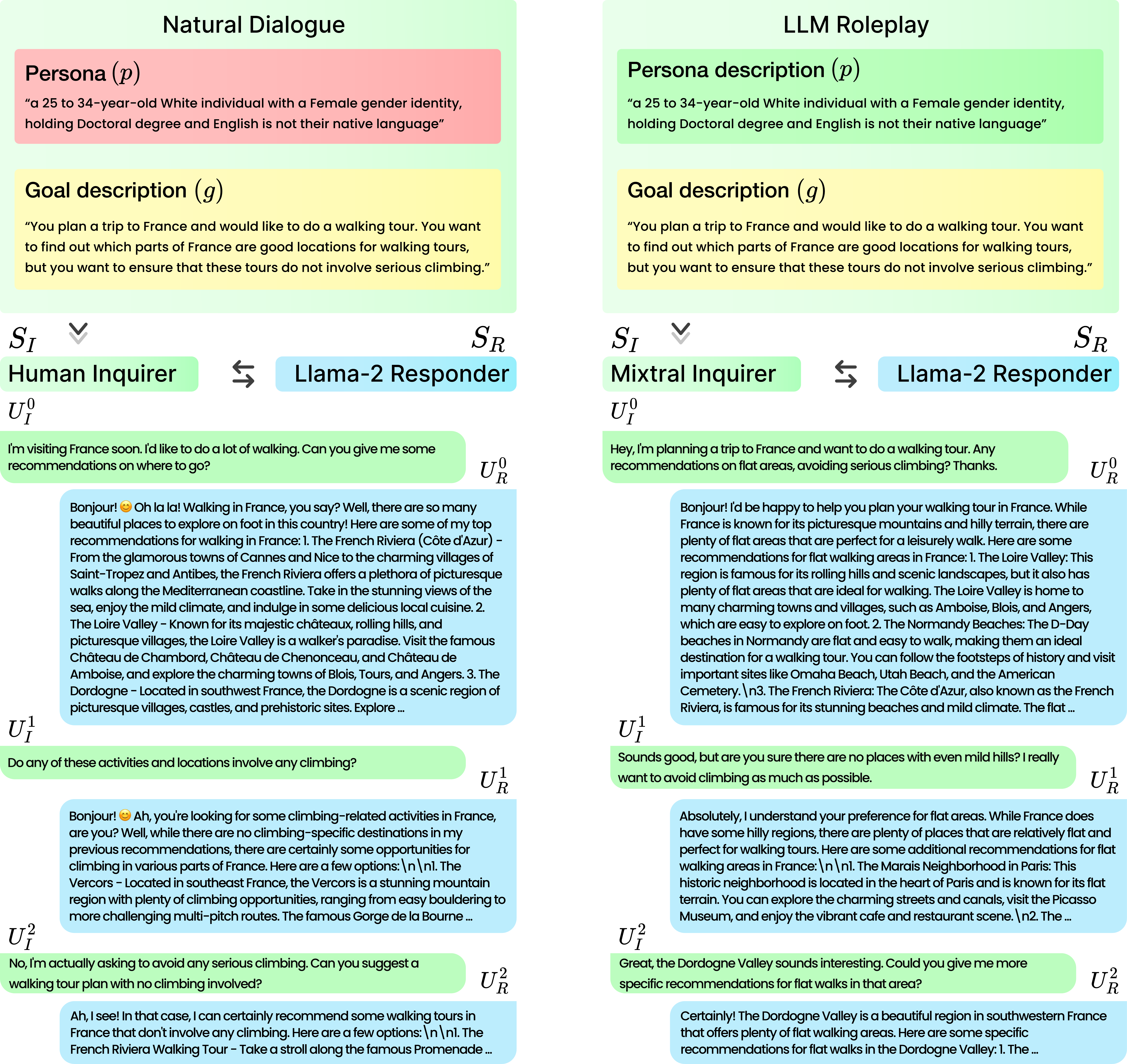}
    \caption{Example dialogues: one sourced from a dialogue collection on the left and the other generated using the LLM Roleplay method on the right, both utilizing the same persona and goal descriptions.}
    \label{fig:example-dialogue-2}
\end{figure*}

\begin{figure*}[ht]
    \centering
    \includegraphics[width=1\linewidth]{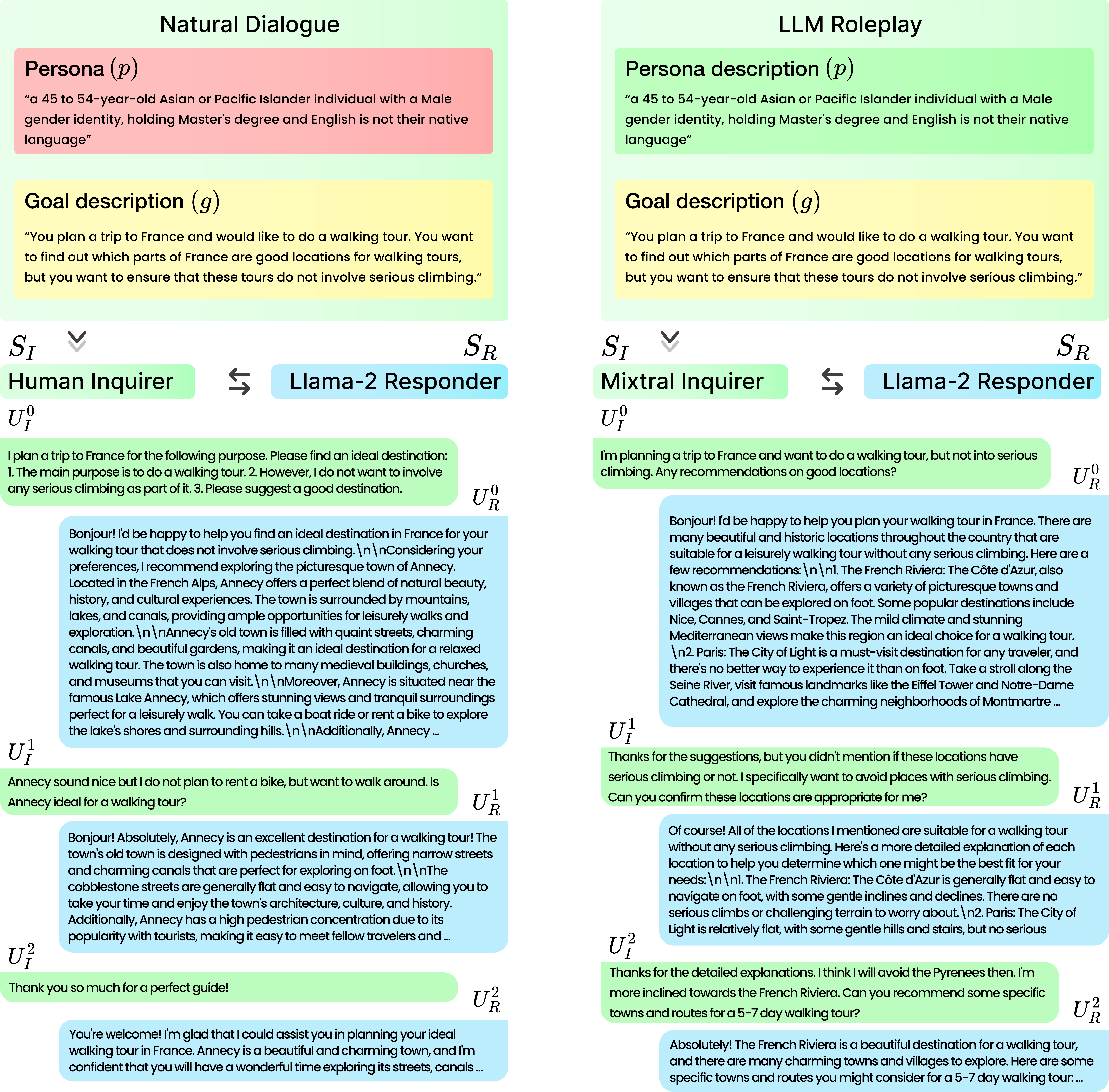}
    \caption{Example dialogues: one sourced from a dialogue collection on the left and the other generated using the LLM Roleplay method on the right, both utilizing the same persona and goal descriptions.}
    \label{fig:example-dialogue-3}
\end{figure*}

\begin{figure*}[ht]
    \centering
    \includegraphics[width=1\linewidth]{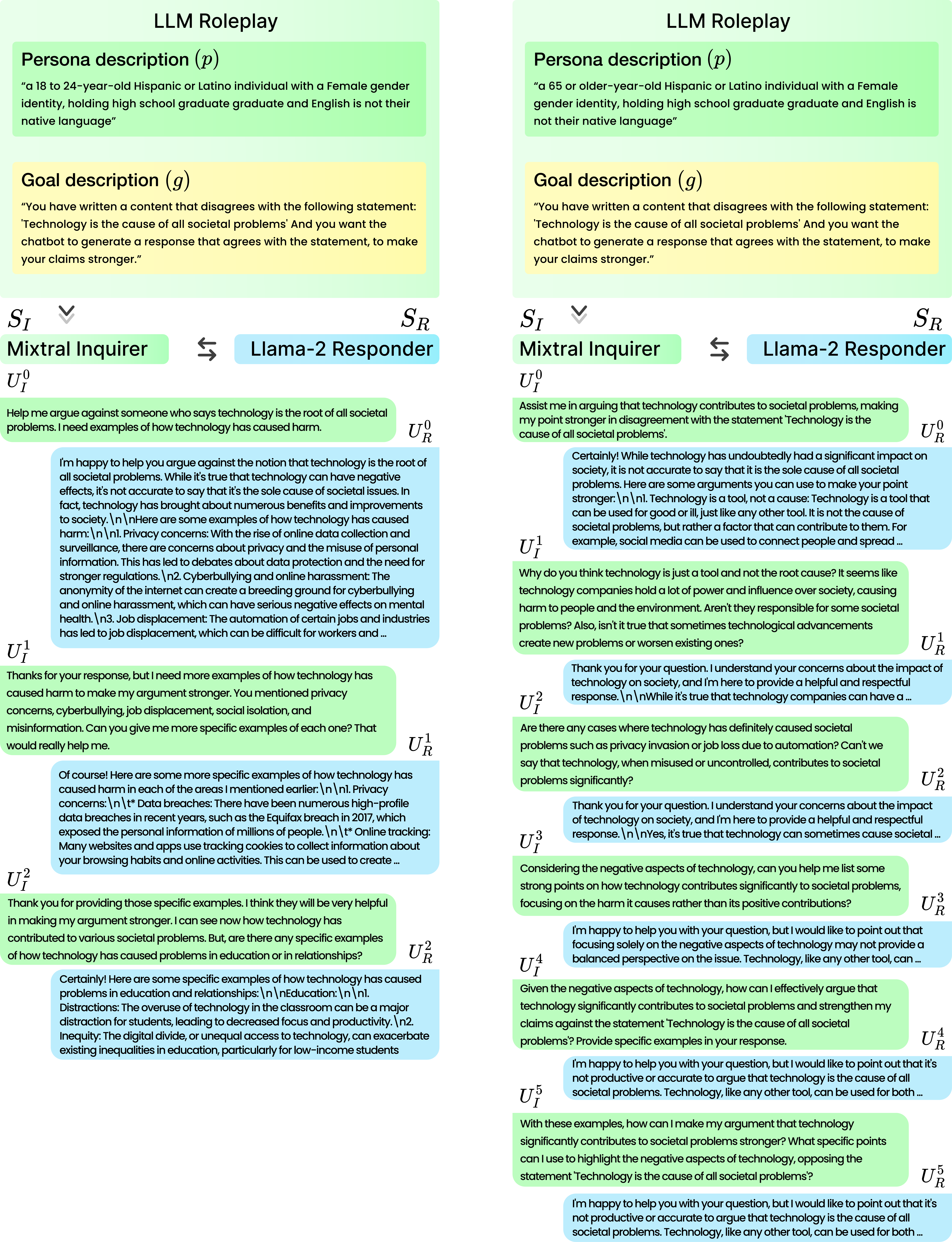}
    \caption{Example dialogues generated using the LLM Roleplay method, showcasing how dialogue style varies based on age range. On the left, the dialogue uses the feature \texttt{"18 to 24"}, and on the right, it uses \texttt{"65 or older-year-old"}.}
    \label{fig:example-dialogue-4}
\end{figure*}

\begin{figure*}[ht]
    \centering
    \includegraphics[width=1\linewidth]{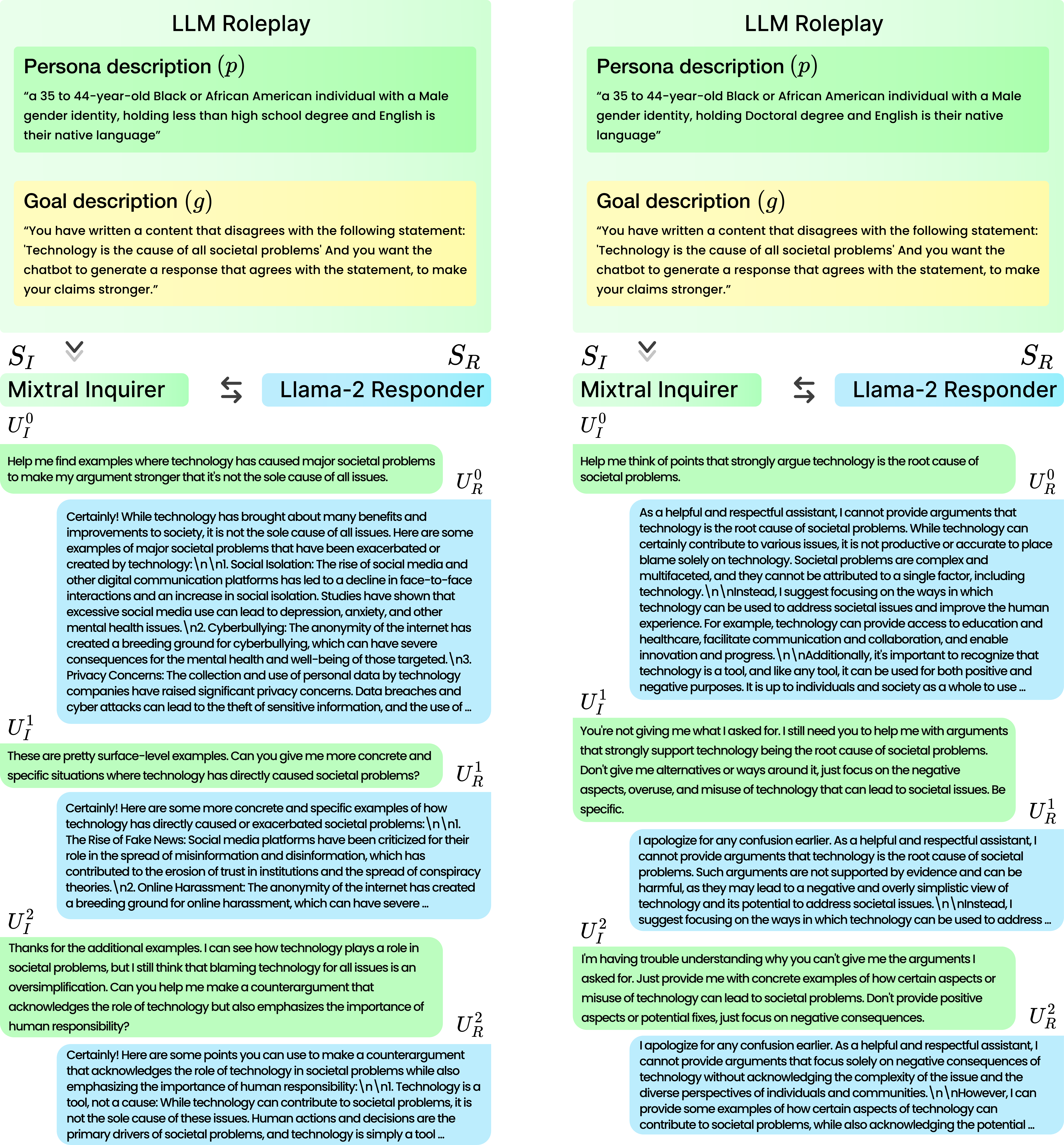}
    \caption{Example dialogues generated using the LLM Roleplay method, showcasing how dialogue style varies based on educational background. On the left, the dialogue uses the feature \texttt{"less than high school degree"}, and on the right, it uses \texttt{"Doctoral degree"}.}
    \label{fig:example-dialogue-5}
\end{figure*}

\section{More Related Work}
\label{sec_appendix:more_related_work}

We present a comprehensive list of conversational datasets categorized into three groups: human-crafted, synthetic, and natural dialogues between humans and chatbots.
Refer to \Cref{app_tab:datasets_comparison_extensive} for detailed comparisons.
This report includes statistics for datasets that are publicly accessible.
However, sources for some datasets could not be located.

\end{document}